\documentclass[sigconf]{acmart}

\usepackage{xspace}
\usepackage{amsfonts}       
\usepackage{amsmath}
\usepackage{nicefrac}       
\usepackage{microtype}      
\usepackage[linesnumbered,lined,boxed]{algorithm2e} 
\usepackage{subcaption}
\usepackage{graphicx}
\usepackage{enumitem}
\usepackage{bbm}
\usepackage{comment}
\usepackage{dsfont}

\AtBeginDocument{%
  \providecommand\BibTeX{{%
    \normalfont B\kern-0.5em{\scshape i\kern-0.25em b}\kern-0.8em\TeX}}}

\newif{\ifhidecomments}
\hidecommentstrue
\ifhidecomments
    \newcommand{\chenhao}[1]{}
    \newcommand{\ram}[1]{}
    \newcommand{\amit}[1]{}
    \newcommand{\divyat}[1]{}
\else
    \newcommand{\chenhao}[1]{\textcolor{blue}{[#1 ---\textsc{ct}]}}
    \newcommand{\ram}[1]{\textcolor{red}{[#1 ---\textsc{ram}]}}
    \newcommand{\amit}[1]{\textcolor{brown}{[#1 ---\textsc{amit}]}}
    \newcommand{\divyat}[1]{\textcolor{orange}{[#1 ---\textsc{divyat}]}}
\fi
\newcommand{\figref}[1]{Fig.~\ref{#1}}

\copyrightyear{2021}
\acmYear{2021}
\setcopyright{acmcopyright}\acmConference[AIES '21]{Proceedings of the 2021 AAAI/ACM Conference on AI, Ethics, and Society}{May 19--21, 2021}{Virtual Event, USA}
\acmBooktitle{Proceedings of the 2021 AAAI/ACM Conference on AI, Ethics, and Society (AIES '21), May 19--21, 2021, Virtual Event, USA}
\acmPrice{15.00}
\acmDOI{10.1145/3461702.3462597}
\acmISBN{978-1-4503-8473-5/21/05}

\newcommand{\para}[1]{\noindent{\bf #1}}

\newcommand{\secref}[1]{\S\ref{#1}}

\newcommand{\vect}[1]{\boldsymbol{#1}}
\newcommand{\vecx}{\vect{x}}
\newcommand{\vecu}{\vect{u}}
\newcommand{\counterfactuals}{\mathcal{C}}
\newcommand{\counterfactual}{\vect{c}}

\newcommand{\classifier}{f}
\DeclareMathOperator*{\argmin}{arg\,min}

\newcommand{\dice}{DiCE\xspace}
\newcommand{\lime}{LIME\xspace}
\newcommand{\shap}{SHAP\xspace}
\newcommand{\dicefa}{$\text{DiCE}_{\text{FA}}$\xspace}
\newcommand{\adult}{{Adult-Income}\xspace}
\newcommand{\german}{{German-Credit}\xspace}
\newcommand{\lclub}{{LendingClub}\xspace}
\newcommand{\triage}{{HospitalTriage}\xspace}
\newcommand{\wachtercf}{WachterCF\xspace}
\newcommand{\wachtershort}{Wachter et al.\xspace}
\newcommand{\wachtercffa}{$\text{WachterCF}_{\text{FA}}$\xspace}

\begin{document}

\title[Towards Unifying Feature Attribution and Counterfactual Explanations]{Towards Unifying Feature Attribution and Counterfactual Explanations: Different Means to the Same End}


\author{Ramaravind Kommiya Mothilal}
\authornote{Work done during stay at Microsoft Research India.}
\affiliation{%
  \institution{Microsoft Research India}
}
\email{raam.arvind93@gmail.com}

\author{Divyat Mahajan}
\affiliation{%
  \institution{Microsoft Research India}
}
\email{t-dimaha@microsoft.com}

\author{Chenhao Tan}
\affiliation{%
  \institution{University of Chicago}
}
\email{chenhao@uchicago.edu}

\author{Amit Sharma}
\affiliation{%
  \institution{Microsoft  Research India}
}
\email{amshar@microsoft.com}


\begin{abstract}
Feature attributions and counterfactual explanations are popular approaches to explain a ML model. 
The former assigns an importance score to each input feature, while the latter provides input examples with minimal changes to alter the model’s predictions.
To unify these approaches, we provide an interpretation based on the actual causality framework and present two key results in terms of their use.
First, we present a method to generate feature attribution explanations from a set of counterfactual examples. These feature attributions convey how important a feature is to changing the classification outcome of a model, especially on whether a subset of features is \textit{necessary} and/or \textit{sufficient} for that change, which attribution-based methods are unable to provide. Second, we show how counterfactual examples can be used to evaluate the goodness of an attribution-based explanation in terms of its necessity and sufficiency. As a result, 
we highlight the complementarity of these two approaches.
Our evaluation on three benchmark datasets --- \adult, \lclub, and \german~--- confirms the complementarity. Feature attribution methods like LIME and SHAP and counterfactual explanation methods like \wachtershort and DiCE often do not agree on feature importance rankings. In addition, by restricting the features that can be modified for generating counterfactual examples, we find that the top-k features from LIME or SHAP are often neither necessary nor sufficient explanations of a model's prediction. Finally, we 
present a case study of different explanation methods on a real-world hospital triage problem. 
\end{abstract}





\maketitle

\section{Introduction}

As complex machine learning (ML) models are being deployed in high-stakes domains like finance and healthcare, explaining why they make a certain prediction has emerged as a critical task. Explanations of a ML model's prediction have found many uses, including to understand the most important features \citep{ribeiro2016should,lundberg2017unified}, discover any unintended bias~\cite{sharma2019certifai}, debug the model~\cite{lakkaraju2016interpretable}, increase trust~\cite{lipton2018mythos,lai+tan:19}, and provide recourse suggestions for unfavorable predictions \cite{wachter2017counterfactual}.

\begin{figure}[t]
    \centering
    \includegraphics[scale=0.35]{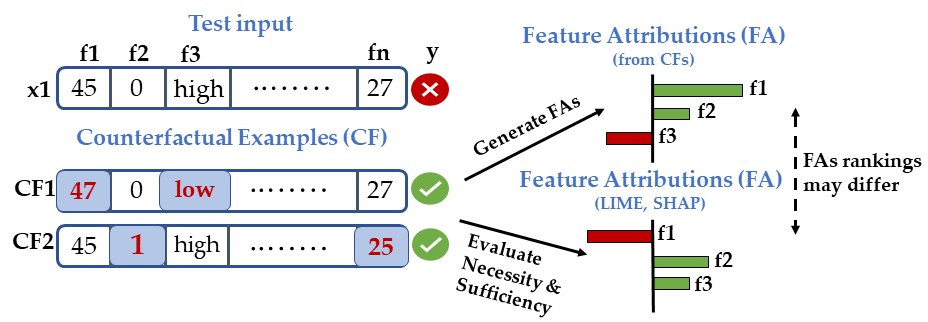}
    \caption{Complementarity of explanation methods.}
    \label{fig:overview}
\end{figure}

There are two popular explanation methods:
\textit{attribution-based} 
and \textit{counterfactual-based}. 
Attribution-based explanations provide a score or ranking over features, conveying the (relative) importance of each feature to the model's output. 
Example methods include local function approximation using linear models~\cite{ribeiro2016should} 
and game-theoretic attribution such as Shapley values~\cite{lundberg2017unified}. The second kind, counterfactual-based explanations, instead generate examples that yield a different model output with minimum changes in the input features, known as  counterfactual examples (CF) \cite{wachter2017counterfactual}. Because of the differences in the type of  output and how they are generated, these two methods are largely studied independent of each other. 


In this paper, we demonstrate the fundamental relationship between attribution-based and counterfactual-based explanations (see Fig.~\ref{fig:overview}).
To provide a formal connection, we introduce the framework of \textit{actual causality}~\cite{halpernbook} to the explanation literature.
Actual causality reasons about the causes of a particular event, while the more common causal inference setting estimates the effect of a particular event~\cite{rubin2005causal,pearl2016causal}. Using actual causality, we define an ideal model explanation and propose two desirable properties for any explanation: \textit{necessity} (is a feature value necessary for generating the model's output?) and \textit{sufficiency} (is the feature value sufficient for generating the model output?).
A good explanation should satisfy both \cite{woodward2006sensitive,lipton1990contrastive}, but we find that current 
explanation 
methods optimize either one of them. CF-based methods like \wachtershort (henceforth ``{\em \wachtercf}'') and \dice~\cite{mothilal2020explaining} find examples that highlight the necessary feature value for a given model output whereas attribution-based methods like \lime~\cite{ribeiro2016should} and \shap~\cite{lundberg2017unified} focus on the sufficiency of a feature value. Thus, the actual causality framework underscores their complementarity:
we need to provide both necessity and sufficiency for a good explanation. 

Our empirical analysis, using \lime and \shap as examples of attribution-based and \wachtercf  and \dice as  examples of  counterfactual-based methods, confirms this complementarity. First, we show that counterfactual-based methods can be used to evaluate explanations from \lime and \shap. By allowing only a specific feature to change in generating CFs, we can evaluate the necessity of the feature's value for the model's predicted output. Similarly, by generating CFs with all but a specific feature, we can evaluate the sufficiency of the feature's value for causing the model's outcome. 
On benchmark datasets 
related to income or credit predictions
(Adult-Income, German-Credit and LendingClub), we find that the top-ranked features from \lime and \shap are often neither necessary nor sufficient. In particular, for Adult-Income and German-Credit, more counterfactuals 
can be generated by using features except the top-3 than using any of the top-3 features, and it is easy to generate counterfactuals even if one of the top-ranked features is not changed at all.   

Second, we show that CF examples can be used to generate feature importance scores that  complement the scores from \lime and \shap. The scores from \dice and \wachtercf do not always agree with those from attribution-based methods: 
\dice and \wachtercf tend to assign relatively higher scores to low-ranked features from \lime and \shap, likely because it is possible to generate valid CFs using those features as well. Ranks generated by the four methods also disagree: not only do attribution-based methods disagree with counterfactual-based methods, but \lime and \shap also disagree on many features and so do \wachtercf and \dice.   

Our results reveal the importance of considering multiple explanation methods to understand the prediction of an ML model. Different methods have different objectives (and empirical approximations). Hence, a single method may not convey the full picture. To demonstrate the value of considering multiple kinds of explanation, we analyze  a high-dimensional real-world dataset that has over 200 features where the ML model's task is to predict whether a patient will be admitted to a hospital. The differences observed above are magnified: \textit{an analyst may reach widely varying conclusion about the ML model depending on which explanation method they choose.} \dice considers triage features as the most important, \lime considers chief-complaint features as the most important, whereas \shap identifies demographic features as the most important. We also find odd results with \lime on necessity: changing the 3rd most important feature provides more valid CFs than changing the most important feature.

To summarize, we make the following contributions:
\begin{itemize}
    \item A unifying framework for attribute-based explanations and counterfactual examples using actual causality;    
    \item A method to evaluate attribution-based methods on the necessity and sufficiency of their top-ranked features; 
    \item Empirical investigation of explanations using commonly used datasets and a high-dimensional dataset. 
\end{itemize}


    

\section{Related Work}
We discuss the desirable properties that any explanation method should have,  the two main types of explanations, and how different explanation methods compare to each other. There is also important work on building intelligible models by design~\cite{caruana2015intelligible,lou2012intelligible,rudin2019stop} that we do not discuss here. 


\subsection{Desirable Properties of an Explanation}
Explanations serve a variety of purposes, including debugging for the model-developer, evaluating properties for an auditor, and providing recourse and trust for an end individual. Therefore, it is natural that explanations have multiple desirable properties based on the context.  \citet{sokol2020explainability} and \citet{miller2018explanation} list the different properties that an explanation ideally should adhere to. Different works have evaluated the soundness~\cite{yeh2019fidelity} (truthfulness to the ML model), completeness~\cite{rawal2020can} (generalizability to other examples), parsimony~\cite{miller2018explanation, dandl2020multi}, and actionability~\cite{ustun2019actionable} of explanations. 
In general, counterfactual-based methods optimize soundness over completeness, while methods that summarize data to produce an attribution score are less sound but optimize for completeness.  

In comparison, the notions of necessity and sufficiency of a feature value for a model's output are less studied.
In natural language processing (NLP), sufficiency and comprehensiveness have been defined based on the output probability in the context of rationale evaluation (e.g., whether a subset of words leads to the same predicted probability as the full text) \cite{yu2019rethinking,deyoung2020eraser,carton2020evaluating}.
By using a formal framework of actual causality \citep{halpernbook}, we define the necessity and sufficiency metrics for explaining any ML model, and provide a method using counterfactual examples to compute them. In concurrent work, Galhotra et al.~\cite{galhotra2021explaining} propose an explanation method based on necessity  and sufficiency metrics. 

\subsection{Attribution-based and Counterfactuals}
Majority of the work in explainable ML provides attribution-based explanations~\cite{sundararajan2017axiomatic, shrikumar2017learning}. Feature attribution methods are local explanation techniques that assign \textit{importance} scores to features based on certain criteria, such as by approximating the local decision boundary~\cite{ribeiro2016should} or estimating the Shapley value~\cite{lundberg2017unified}. A feature's score 
captures
its contribution to the predicted value of an instance. 
In contrast, counterfactual explanations~\cite{wachter2017counterfactual, poyiadzi2020face, dhurandhar2018explanations,karimi2020model,dandl2020moc,ustun2019actionable,verma2020counterfactual} are minimally-tweaked versions of the original input that lead to a different predicted outcome than the original prediction. 
In addition to proximity to the original input, it is important to ensure feasibility~\cite{karimi2020algorithmic}, real-time response \cite{schleich2021geco}, and diversity among counterfactuals~\cite{russell2019efficient,mothilal2020explaining}. 

We provide a unified view of these two explanations. They need not be considered separate (Research Challenge 1 in Verma et al.~\cite{verma2020counterfactual}): counterfactuals can provide  another way to generate feature attributions, as suggested by \citet{sharma2020certifai} and \citet{barocas2020hiddencf}. We extend this intuition by conducting an extensive empirical study on the attributions generated by counterfactuals, and comparing them to other attribution-based methods. In addition, we introduce a formal causality framework to show how different explanation methods correspond to different notions of a feature ``causing'' the model output: counterfactuals focus on the necessity of a feature while other methods tend to focus on its sufficiency to cause the model output.  



\section{Actual Causality: Unifying explanations} \label{sec:causal}

Let $\classifier(\vecx)$ be a machine learning model and $\vecx$ denote a vector of $d$ features, $(x_1, x_2, ...x_d)$. Given input $\vecx_0$ and the output $\classifier(\vecx_0)$, a common explanation task is to determine which features are responsible for this particular prediction.  

Though both attribution-based and counterfactual-based methods aim to explain a model's output at a given input, the difference and similarity in their implications are not clear. 
While feature attributions highlight features that are important in terms of their contributions to the model prediction, it does not imply that changing important features is sufficient or necessary to lead to a different (desired) outcome.
Similarly, while CF explanations provide 
insights 
for reaching a different outcome, the features changed may not include the most important features of feature attribution methods.

Below we show that while these explanation methods may appear distinct, they are all motivated by the same principle of whether a feature is a ``cause'' of the model's prediction, and to what extent. 
We provide a formal framework based on actual causality~\cite{halpernbook} to interpret them.

\subsection{Background: Actual Cause and Explanation}
We first define \textit{actual cause} and how it can be used to explain an event. In our case, the classifier's prediction is an event, and the input features are the potential causes of the event.    
According to \citet{halpernbook}, causes of an event are defined w.r.t to a structural causal model (SCM) that defines the relationship between the potential causes and the event. In our case, the learnt ML model $\classifier$ is the SCM ($M$) that governs how the prediction output is generated from the input features. The structure of the SCM consists  of each feature as a node that causes other intermediate nodes (e.g., different layers of a neural network), and then finally leads to the output node. We assume that the feature values are generated from an unknown process governed by a set of parameters that we collectively denote as $\vecu$, or the \emph{context}. Together, $(M, \vecu)$  define a specific configuration of the input $\vecx$ and the output $\classifier(\vecx)$ of the model. 

For simplicity, the following definitions assume that individual features are independent of each other, and thus any feature can be changed without changing other features. However, in explanation goals such as algorithmic recourse it is important to consider the causal dependencies between features themselves~\cite{mahajan2019preserving,karimi2020algorithmic,kumar2020problems, galhotra2021explaining}; we leave such considerations for future work. 

\begin{definition}[\textbf{Actual Cause, (Original definition)}~\cite{halpernbook}] \label{def:actual-cause}
A subset of feature values  $\vecx_{j}=a$ is an actual cause of the model output $\classifier(\vecx_{-j}=b, \vecx_{j}=a)=y^*$ under the causal setting $(M, \vecu)$ if all the following conditions hold: 
\begin{enumerate}
    \item Given $(M, \vecu)$, $\vecx_j=a$ and $\classifier(\vecx_{-j}=b, \vecx_j=a)=y^*$.
    \item There exists a subset of features $W \subseteq \vecx_{-j}$ such that if $W$ is set to $w'$, then $(\vecx_j \leftarrow a, W \leftarrow w') \Rightarrow (y=y^*)$ and  $(\vecx_j \leftarrow a',  W \leftarrow w') \Rightarrow y \neq y^*$ for some value $a'$.
    \item $\vecx_j$ is minimal, namely, there is no strict subset $\vecx_s \subset \vecx_j$ such that $\vecx_s=a_s$ satisfies conditions 1 and 2, where $a_s \subset a$.
\end{enumerate}
\end{definition}
In the notation above, $\vecx_i \leftarrow v$ denotes that $\vecx_i$ is intervened on and set to the value $v$, irrespective of its observed value under $(M, \vecu)$.  Intuitively, a subset of feature values $\vecx_j=a$ is an actual cause of $y^*$ if under some value $b'$  of the other features $\vecx_{-j}$, there exists a value  $a'\neq a$ such that  $\classifier(\vecx_{-j}=b', a') \neq y^*$ and  $\classifier(\vecx_{-j}=b', a) = y^*$.
For instance, consider  a linear model with three binary features $\classifier(x_1, x_2, x_3)= I(0.4x_1 + 0.1x_2 + 0.1x_3 >= 0.5)$ and an observed prediction of $y=1$. Here each feature $x_i=1$ can be considered an actual cause for the model's output, since there is a context where its value is needed to lead to the outcome $y=1$. 

To differentiate between the contributions of features, we can use a stronger definition, the \emph{but-for} cause. 

\begin{definition}[\textbf{But-for Cause}]
A subset of feature values $\vecx_{j}=a$ is a but-for cause of the model output $\classifier(\vecx_{-j}=b, \vecx_{j}=a)=y^*$ under the causal setting $(M,\vecu)$ if it is an actual cause and the empty set $W=\phi$ satisfies condition 2. 
\end{definition}
That is, changing the value of $x_j$ alone changes the prediction of the model at $\vecx_0$. On the linear model, now we obtain a better picture: $x_1=1$ is always a but-for cause for $y=1$. 
The only context in which  $x_2=1$ and $x_3=1$ are but-for causes for $y=1$ is when $x_1=1$. 

While the notion of but-for causes captures the \textit{necessity} of a particular feature subset for the obtained model output, it does not capture \textit{sufficiency}. Sufficiency means that setting a feature subset $\vecx_j \leftarrow a$ will always lead to the given model output, irrespective of the values of other features. To capture sufficiency, therefore, we need an additional condition.
{
\begin{equation} \label{eq:suff-cause}
\vecx_j \leftarrow a \Rightarrow y=y^* \text{\ \ } \forall \vecu \in U 
\end{equation}%
}%
That is, for the feature subset value $\vecx_j=a$ to be a sufficient cause,  the above statement should be valid in \textit{all} possible contexts. 
Based on the above definitions, we are now ready to define an ideal explanation that combines the idea of actual cause and sufficiency.
\begin{definition}[\textbf{Ideal Model Explanation}]
A subset of feature values $\vecx_j=a$ is an explanation for a model output $y^*$ relative to {\em a set of contexts} $U$, if 
\begin{enumerate}
\item \textbf{Existence:} There exists a context $\vecu \in U$ such that  $\vecx_j=a$ and $f(\vecx_{-j}=b, \vecx_j=a)=y^*$.
\item \textbf{Necessity:} For each context $\vecu \in U$ where $\vecx_j=a$ and $f(\vecx_{-j}=b, \vecx_j=a)=y^*$, some feature subset $\vecx_{sub} \subseteq \vecx_j$ is an actual cause under $(M, \vecu)$ (satisfies conditions 1-3 from Definition~\ref{def:actual-cause}).
    \item \textbf{Sufficiency:} For all contexts $\vecu' \in U$, $\vecx_j \leftarrow a \Rightarrow y=y^*$.
    \item \textbf{Minimality:} $\vecx_j$ is minimal, namely, there is no strict subset $\vecx_s \subset \vecx_j$ such that $\vecx_s=a_s$ satisfies conditions 1-3 above, where $a_s \subset a$.
\end{enumerate}

\end{definition}

This definition captures the intuitive meaning of explanation. For a given feature $x$,  condition 2 states that the feature affects the output (output changes if the feature is changed under certain conditions), and condition 3 states that as long as the feature is unchanged, the output cannot be changed. 
In practice, however, it is rare to find such clean explanations of a ML model's output. Even in our simple linear model  above, no feature is sufficient to cause the output, $y=1$. 

\subsection{Partial Explanation for Model Output}
For most realistic ML models, an ideal explanation is impractical.  Therefore, we now describe the concept of  
\emph{partial} explanations~\cite{halpernbook} that relaxes the necessity and sufficiency conditions to consider the fraction of contexts over which these conditions are valid. 
Partial explanations are characterized by two metrics.

The first metric captures the extent to which a subset of feature values is \textit{necessary} to cause the model's (original) output.
{
\begin{equation}\label{eq:orig-alpha}
    \alpha = \Pr(x_{j} \text{ is a cause of } y^*|\vecx_j=a, y=y^*) 
\end{equation}%
}%
where `is a cause' means that $x_j=a$ satisfies Definition 3.1.
The second metric captures \textit{sufficiency} using conditional probability of outcome given the subset of feature values. 
{
\begin{equation} \label{eq:beta}
    \beta = \Pr(y=y^*|\vecx_j \leftarrow a) 
\end{equation}%
}%
where $\vecx_j \leftarrow a$ denotes an intervention to set $\vecx_j$ to $a$. 
Both probabilities are over the set of contexts $U$. 
Combined, they can be called $(\alpha, \beta)$ goodness of an explanation. When both $\alpha=1$ and $\beta=1$, $\alpha=1$ captures that $\vecx_{j}=a$ is a necessary cause of $y=y^*$ and $\beta=1$ captures that $\vecx_{j}=a$ is a sufficient cause of $y=y^*$.
In other words,  a subset of feature values $x_j=a$ is a good explanation for a model's output $y^*$ if it is an actual cause of the outcome and $y=y^*$ with high probability whenever $x_j=a$. 

\subsection{Unifying Different Local Explanations}
Armed with the $(\alpha, \beta)$ goodness of explanation metrics, we now  show how common explanation methods can be considered as special cases of the above framework. 

\para{Counterfactual-based explanations.}
First, we show how counterfactual explanations relate to $(\alpha, \beta)$:
When only but-for causes (instead of actual causes) are allowed,  
$\alpha$ and $\beta$ capture the intuition behind counterfactuals.
Given $y=y^*$ and a candidate feature subset $\vecx_j$,  $\alpha$  corresponds to fraction of contexts where $x_{j}$ is a but-for cause. That is, keeping everything else constant and only changing $\vecx_{j}$, how often does the classifier's outcome change? Eqn.~\ref{eq:orig-alpha} reduces to,
{
\begin{equation} \label{eq:alpha-cf}
    \alpha_{CF} = \Pr( (\vecx_j \leftarrow a' \Rightarrow y \neq y^*)| \vecx_j=a, \vecx_{-j} = b, y=y^*)
\end{equation}%
}%
\noindent where the above probability is over a reasonable set of contexts (e.g., all possible values for discrete features and a bounded region around the original feature value for continuous features). By definition, each of the perturbed inputs above that change the value of $y$ can be considered as a counterfactual example~\cite{wachter2017counterfactual}. Counterfactual explanation methods aim to find the smallest perturbation in the feature values that change the output, and correspondingly the modified feature subset $x_j$ is a but-for cause of the output. 
$\alpha_{CF}$ provides a metric to summarize the outcomes of all such perturbations and  to rank any feature subset for their necessity in generating the original model output. In practice, however, computing $\alpha$ is computationally prohibitive and therefore explanation methods empirically find a set of counterfactual examples and allow (manual) analysis on the found counterfactuals. In \secref{sec:methods}, we will see how we can develop a feature importance score using counterfactuals that is inspired from the $\alpha_{CF}$ formulation. 


$\beta$ corresponds to the fraction of contexts where $x_j =a$ is sufficient to keep $y=y^*$. That corresponds to the degree of sufficiency of the feature subset: keep $\vecx_j$ constant but change everything else and check how often the outcome remains the same. While not common, such  a perturbation can be considered as a special case of the counterfactual generation process, where we specifically restrict change in the given feature set.  A similar idea is explored in (local) anchor explanations \cite{ribeiro2018anchors}. It is also related to pertinent positives and pertinent negatives \citep{dhurandhar2018explanations}. 

\para{Attribution-based explanations.}
Next, we show the connection of attribution-based explanations with $(\alpha, \beta)$.
$\beta$ is defined as in Eqn.~\ref{eq:beta}, the fraction of all contexts where $\vecx_j \leftarrow a$ leads to $y=y^*$.
Depending on how we define the set of \emph{all} contexts, we obtain different local attribute-based explanations. The total number of contexts is $2^m$ for $m$ binary features and is infinite for continuous features.
For ease of exposition, we consider binary features below. 

LIME can be interpreted as estimating $\beta$ for a restricted set of contexts (random samples) near the input point. Rather than checking Eqn.~\ref{eq:suff-cause} for each of the random sampled points and estimating $\beta$ using Eqn.~\ref{eq:beta}, it uses linear regression to estimate $\beta(a, y^*)- \beta(a', y^*)$. Note that linear regression estimates 
$\mathbb{E}[Y|\vecx_j=a]- \mathbb{E}[Y|\vecx_j=a']$ are equivalent to  $\Pr[Y=1|\vecx_j=a]-\Pr[Y=1|\vecx_j=a']$ for a binary $y$. 
LIME estimates effects for all features at once using linear regression, assuming that each feature's importance is independent. 

Shapley value-based methods take a different approach.
Shapley value for a feature is defined as the number  of times that including a feature leads to the observed outcome, averaged over all possible configurations of other input features. That is, they define the valid contexts for a feature value as all valid configurations of the other features  (size $2^{m-1}$). The intuition is to see, at different values of other features, whether the given feature value is sufficient to cause the desired model output $y^*$.  The goal of estimating Shapley values corresponds to the equation for $\beta$ described above (with an additional term for comparing it to the base value).  


Note how selection of the contexts effectively defines the type of attribution-based  explanation method~\cite{sundararajan2019many,kumar2020problems}. For example, we may weigh the contexts based on their likelihood 
in some world model,
leading to \textit{feasible} attribute 
explanations~\cite{aas2019explaining}. 

\para{Example and practical implications.} The above analysis indicates that different explanation methods optimize for either $\alpha$ or $\beta$: counterfactual explanations are inspired from the $\alpha_{CF}$ metric
and attribution-based methods like LIME and SHAP  from the $\beta$ metric. Since $\beta$ focuses on the power of a feature to lead to the observed outcome and $\alpha$ on its power to change the outcome conditional  that the (feature, outcome) are already observed, the two metrics need not be the same. 
 For example, consider a model, $y=I(0.45x_1+0.1x_2\geq 0.5)$ where $x_1,x_2 \in [0,1]$ are  continuous features,  and an input point $(x_1=1, x_2=1, y=1)$. To explain this prediction, LIME or SHAP will assign high importance to $x_1$ compared to $x_2$ since it has a higher coefficient value of 0.45. Counterfactuals would also give importance to $x_1$ (e.g., reduce $x_1$ by 0.12 to obtain $y=0$), but also suggest to change $x_2$ (e.g., reduce $x_2$ to 0.49), depending on how the loss function from the original input is defined  (which defines the set of contexts for $\alpha$). Suppl. \ref{app:exp-simple} shows the importance scores by different methods for this example.

Therefore, a good explanation ideally needs both high $\alpha$ and $\beta$ to provide the two different facets. Our framework suggests that there is value in 
evaluating both qualities for an explanation method, and in general considering both types of explanations for their complementary value in  understanding  a model's output. In the following, we propose methods for evaluating necessity ($\alpha_{CF}$) and sufficiency ($\beta$) of an explanation and study their implications 
in real-world  datasets. 

\section{PROPOSED METHODS} 
\label{sec:methods}

To connect attribution-based methods with counterfactual explanation, we propose two methods. The first measures the necessity and sufficiency of any attribution-based  explanation  using counterfactuals, and the second creates feature importance scores using counterfactual examples.  

\subsection{Background: Explanation methods}
For our empirical evaluation, we looked for explanation methods that are publicly available on GitHub. For attribution-based methods, we use the two most popular open-source libraries,  LIME~\cite{ribeiro2016should} and SHAP~\cite{lundberg2017unified}. We choose counterfactual methods based on their popularity and whether a method supports generating CFs using user-specified feature subsets (a requirement for our experiments).  Alibi~\cite{alibi}, AIX360~\cite{arya2019one}, DiCE~\cite{mothilal2020explaining}, and MACE~\cite{karimi2020model} are most popular on GitHub, but only DiCE explicitly supports CFs from feature subsets (more details about method selection are in Suppl. \ref{app:cf-algos}). We also implemented the seminal method from \wachtershort for CF explanations, calling it \wachtercf.

\para{Attribution-based methods.} 
For a given test instance $\vecx$ and a ML model $f(.)$, LIME perturbs its feature values and uses the perturbed samples to build a local linear model $g$ of complexity $\Omega(g)$. 
The coefficients of the linear model are used as explanations $\zeta$ and larger coefficients imply higher importance. Formally, \lime generates explanations by optimizing the following loss where $L$ measures how close $g$ is in approximating $f$ in the neighborhood of $\vecx$, $\pi_{\vecx}$.
{
\begin{align}
    \zeta(\vecx) = \argmin_{g \in G} \operatorname{L}(\classifier,g,\pi_{\vecx}) +
    \Omega(g)
    \label{eq:lime}
\end{align}
}
\shap, on the other hand, assigns importance score to a feature based on Shapley values, which are computed using that feature's average marginal contribution across different coalitions of all features. 

\para{Counterfactual generation method.} For counterfactual explanations, the method from Wachter et al. optimizes the following loss, where $\counterfactual$ is a counterfactual example.
{
\begin{equation}
    \counterfactual^* = \argmin_{\counterfactual}  \operatorname{yloss}(\classifier(\counterfactual), y) +
                                  \lambda_1  dist(\counterfactual, \vecx)  
\label{eq:wachtercf}
\end{equation}
}
The two additive terms in the loss minimize  (1) $\operatorname{yloss}(.)$ between ML model $f(.)$'s prediction and the desired outcome $y$, (2) distance between $\counterfactual_i$ and test instance $\vecx$. For obtaining multiple CFs for the same input, we simply re-initialize the optimization with a new random seed.
As a result, this method may not be able to find unique CFs.

The second method, DiCE, handles the issue of multiple unique CFs by introducing a diversity term to the loss, using a determinantal point processes based method~\cite{kulesza2012determinantal}.  It returns a diverse set of $nCFs$ counterfactuals by solving a combined  optimization problem over multiple CFs, 
where $\counterfactual_i$ is a counterfactual example:
{
\begin{align}
    \counterfactuals(\vecx) = \argmin_{\counterfactual_1, \ldots, \counterfactual_{nCF}} &  \frac{1}{nCF} \sum_{i=1}^{nCF} \operatorname{yloss}(\classifier(\counterfactual_i), y) +
                                  \frac{\lambda_1}{nCF} \sum_{i=1}^{nCF} dist(\counterfactual_i, \vecx)  \nonumber \\ & - \lambda_2 \operatorname{dpp\_diversity}(\counterfactual_1, \ldots, \counterfactual_{nCF}). \label{eq:dice}
\end{align}
}

\subsection{Measuring Necessity and Sufficiency} \label{sec:necc-suff}
Suppose $y^*=f(\vecx_j=a, \vecx_{-j}=b)$ is the output of a classifier $f$ for input $\vecx$. To measure necessity of a feature value $\vecx_j=a$ for the model output $y^*$, we would like to operationalize Equation~\ref{eq:alpha-cf}. A simple way 
is to use a method for generating counterfactual explanations, but restrict it such that only $\vecx_j$ can be changed. 
The fraction of times that changing $\vecx_j$  leads to a valid counterfactual example indicates that the extent to which $\vecx_j=a$ is necessary for the current model output $y^*$. That is, if we can change the model's output by changing $\vecx_j$, it means that the $\vecx_j$ features' values are necessary to generate the model's original output. Necessity is thus defined as
{
\begin{equation}
    \operatorname{Necessity} = \frac{\sum_{i, \vecx_j \neq a} \mathds{1}  (CF_{i})}{\text{nCF}*N},
\end{equation}
}where $N$ is the total number of test instances for which \textit{nCF} counterfactuals are generated each. 

For the sufficiency condition from Equation~\ref{eq:beta}, we adopt the reverse approach. Rather than changing $\vecx_j$, we fix it to its original value and let all other features vary their values,  
If no unique valid counterfactual examples are generated, then it implies that $\vecx_j=a$ is sufficient for causing the model output $y^*$. If not, then (1- fraction of times that unique CFs are generated) tells us about the extent of sufficiency of $\vecx_j=a$. In practice, even when using all the features, we may not obtain 100\% success in generating valid counterfactuals. Therefore, we modify the sufficiency metric to compare the fraction of unique CFs generated using all features to the fraction of unique CFs generated while keeping $\vecx_j$ constant (in other words, we encode the benchmark of using all features to generate CFs in the definition of sufficiency): 
{
\begin{equation}
    \operatorname{Sufficiency} = \frac{\sum_{i} \mathds{1}  (CF_{i})}{\text{nCF}*N} - \frac{\sum_{i, \vecx_j \leftarrow a} \mathds{1}  (CF_{i})}{\text{nCF}*N}
\end{equation}
}

\subsection{Feature Importance using Counterfactuals}
In addition to evaluating properties of attribution-based explainers,  counterfactual explanations offer a natural way of generating feature attribution scores based on the extent to which a feature value is necessary for the outcome. The intuition comes from Equation~\ref{eq:alpha-cf}:  a feature that is changed more often when generating counterfactual examples must be an important feature. Below we describe the methods, \wachtercffa and \dicefa to generate attribution scores from a set of counterfactual examples.



To explain the output $y^*=\classifier(\vecx)$, the \dicefa algorithm proceeds by generating a diverse set of $\text{nCF}$ counterfactual examples for the input $\vecx$, where $\text{nCF}$ is the number of CFs. To generate multiple CFs using \wachtercf, we run the optimization in Eqn.~\ref{eq:wachtercf} multiple times with random initialization as suggested by Wachter et al.
A feature $x_j$ that is important in changing a predicted outcome, is more likely to be changed frequently in $\text{nCF}$ CFs than a feature $x_k$ that is less important.  For each feature, therefore, the attribution score is the fraction of CF examples that have a modified value of the feature. To generate a local explanation, the attribution score is averaged over multiple values of $\text{nCF}$, typically going from 1 to 8. 
To obtain a global explanation, this attribution score is averaged over many test inputs.

\subsection{Datasets and Implementation Details} \label{sec:data}
We use three common datasets in explainable ML literature: \adult~\cite{adult}, \lclub~\cite{tan2017detecting}, \german~\cite{german}. We use the default hyperparameters for LIME, SHAP (using KernelExplainer) and DiCE. For the counterfactual methods, we use the same value of $\lambda_1$ (0.5) for both \dice (Eqn.~\ref{eq:dice}) and \wachtercf (Eqn.~\ref{eq:wachtercf}) and set $\lambda_2$ to 1.0. The results presented are robust to different choices of hyperparameters of proximity and diversity (see Suppl. \ref{app:valid-stable}). 
More details about the dataset and implementation are in the Suppl. ~\ref{app:impl-details}.

\begin{figure*}[!ht]
\centering
\begin{subfigure}{.48\textwidth}
  \centering
  \includegraphics[width=0.9\textwidth]{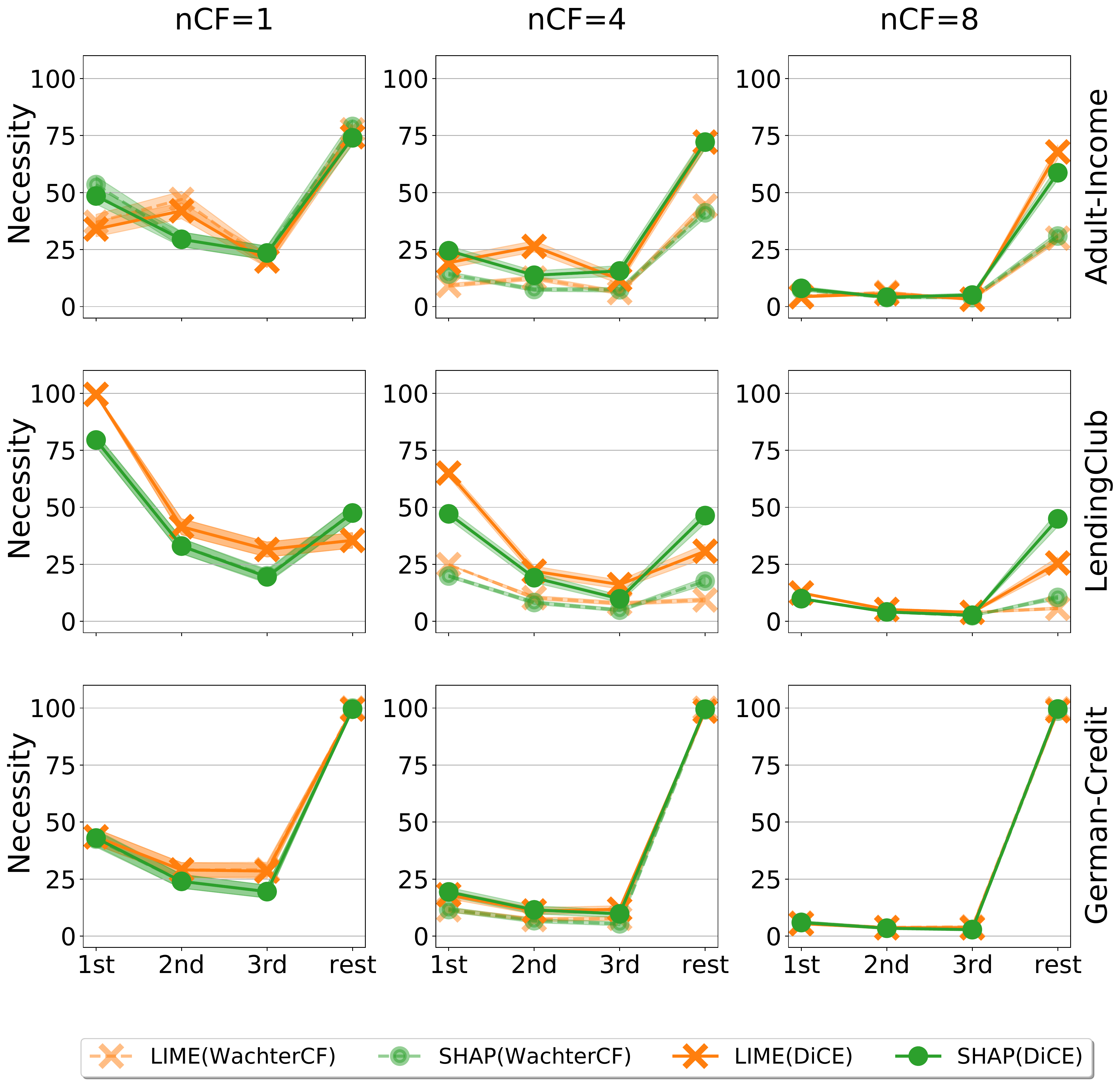}
  \caption{Necessity}
  \label{fig:all_data_gen_cfs_topk_feats}
\end{subfigure}
\hfill
\begin{subfigure}{.48\textwidth}
  \centering
  \includegraphics[width=0.9\textwidth]{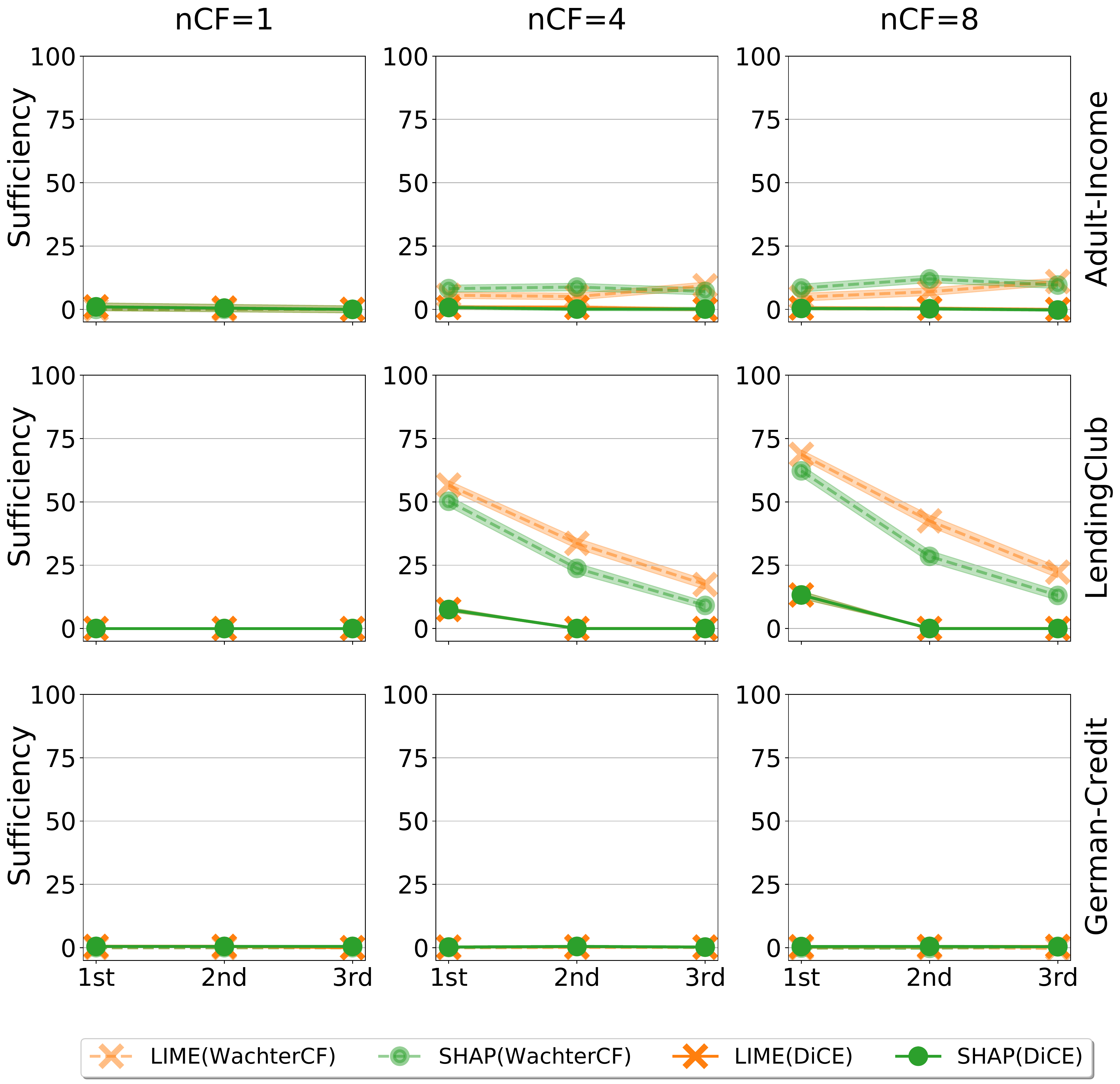} 
  \caption{Sufficiency}
  \label{fig:all_data_gen_cfs_fix_topk_feats}
\end{subfigure}%
\caption{The $y$-axis represents the \textit{necessity} and \textit{sufficiency} measures at a particular $\text{nCF}$, as defined in \secref{sec:necc-suff}.
In \figref{fig:all_data_gen_cfs_topk_feats}, we are only allowed to change the $k$-th most important features ($k=1,2,3$) or the other features,
whereas in \figref{fig:all_data_gen_cfs_fix_topk_feats}, we fix the $k$-th most important features ($k=1,2,3$) but are allowed to change other features.
While necessity is generally aligned with feature ranking derived from LIME/SHAP, 
the most important features often cannot lead to changes in the model output on their own.
In almost all cases, ``rest'' achieves better success in producing CFs using both \dice and \wachtercf.
For sufficiency, none of these top features are sufficient to preserve original model output. \dice and \wachtercf differ the most for \lclub with $nCF>1$, where latter's difficulty to generate unique multiple CFs increases the measured sufficiency of a feature.}
\end{figure*}

\section{Evaluating Necessity \& Sufficiency} \label{sec:topk-feats}


We start by examining the necessity and sufficiency of top features derived with feature attribution methods through counterfactual generation.
Namely, we measure whether we can generate valid CFs by changing only the $k$-th most important feature (necessity) or changing other features except the $k$-th most important feature (sufficiency).
Remember that necessity and sufficiency are defined with respect to the original output.
For example, if changing a feature can vary the predicted outcome, then it means that this feature is necessary for the original prediction.

\para{Are important features necessary?}
Given top features identified based on feature attribution methods (\lime and \shap), we investigate whether we can change the prediction outcomes by using \textit{only} the $k$-th most important feature, where $k$ $\in$ \{1,2,3\}, We choose small $k$ since the number of features is small in these datasets.
Specifically, we measure the average percentage of \textit{unique} and \textit{valid} counterfactuals generated using \dice and \wachtercf for 200 random test instances by fixing other features and changing only the $k$-th most important feature.
This analysis helps us understand if the top features from \lime or \shap are \textit{necessary} to produce the current model output.
\figref{fig:all_data_gen_cfs_topk_feats}
shows the results for different datasets when asked to generate different numbers of CFs. While we produced CFs for $\text{nCF}$ $\in$ \{1,2,4,6,8\}, we show results only for 1, 4, and 8 for brevity.
To provide a benchmark, we also consider the case where we use all the other features that are not in the top three.

Our results in Fig.~\ref{fig:all_data_gen_cfs_topk_feats} suggest that the top features are mostly unnecessary for the original prediction: changing them is less likely to alter the predicted outcome.
For instance, in \german, none of the top features have a necessity of above 50\%, in fact often below 30\%.
In comparison, features outside the top three can always achieve almost 100\%.
This is likely related to the fact that there are 20 features in \german, but the observation highlights the {\em limited} utility in explanation by focusing on the top features from feature attribution methods.
Similar results also show up in \adult, but not as salient as in \german.

In \lclub, we do find that the top feature is relatively higher on the necessity metric. 
Upon investigation, we find this dataset has a categorical feature \textit{grade} of seven levels, which is 
assigned by the lending company as an indicator of loan repayment. The loan grade is designed based on a  combination of factors including credit score. Since the quality of loan grade is highly correlated with loan repayment status, both \lime and \shap give high importance score to this feature for most test instances -- they assign highest score for 98\% and 73\% of the test instances respectively.  
As a result, changing \lime's top-1 feature is enough to get almost perfect unique valid CFs when generating one counterfactual.
However, the necessity of a single feature quickly reduces as we generate more CFs.
Even in this dataset where there is a dominant feature, the features other than the top-3 become more necessary than the top feature (grade) for $nCF>4$ and when diversity is enforced using \dice. 


That said, necessity is generally aligned with the feature ranking from \lime and \shap: the higher the feature importance score, the greater the necessity.
The only exception is the second most important feature in \adult based on \lime.
For most instances, this feature is a person's education level. 

We repeat the above analysis by allowing \textit{all} features upto top-$k$ to be changed (details in Suppl. \ref{sec:app-suff-nec}) and find that necessity of the top-$k$ subset increases, but is still less than 100\% for nCF$>1$. That is, changing all top-3 ranked features is also not enough to generate CFs for all input examples, especially for higher-dimensional \german.

\begin{figure*}[t]
    \centering
    \begin{subfigure}{.48\textwidth}
    \centering
    \includegraphics[width=.9\textwidth]{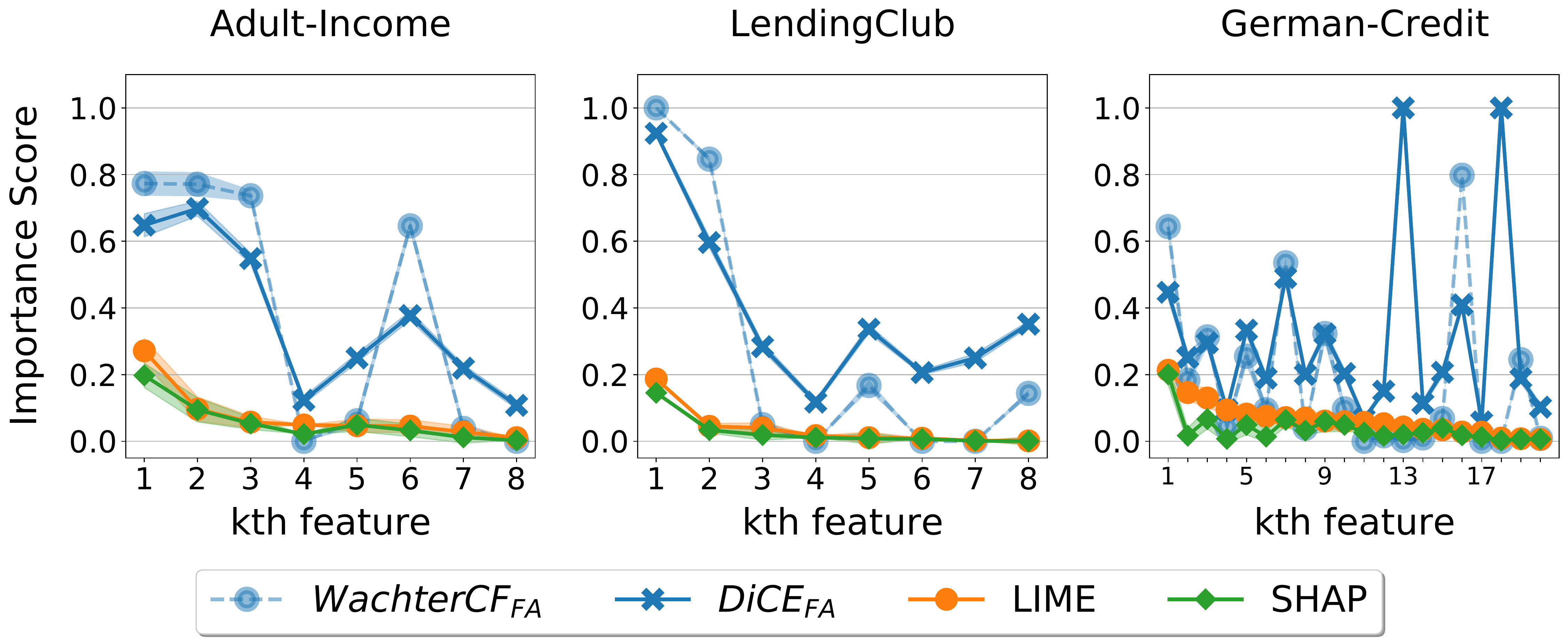}
    \caption{Average feature importance scores (nCF=4).}
    \label{fig:feature_attribution_scores}
    \end{subfigure}
    \begin{subfigure}{.48\textwidth}
    \centering
    \includegraphics[width=.9\textwidth]{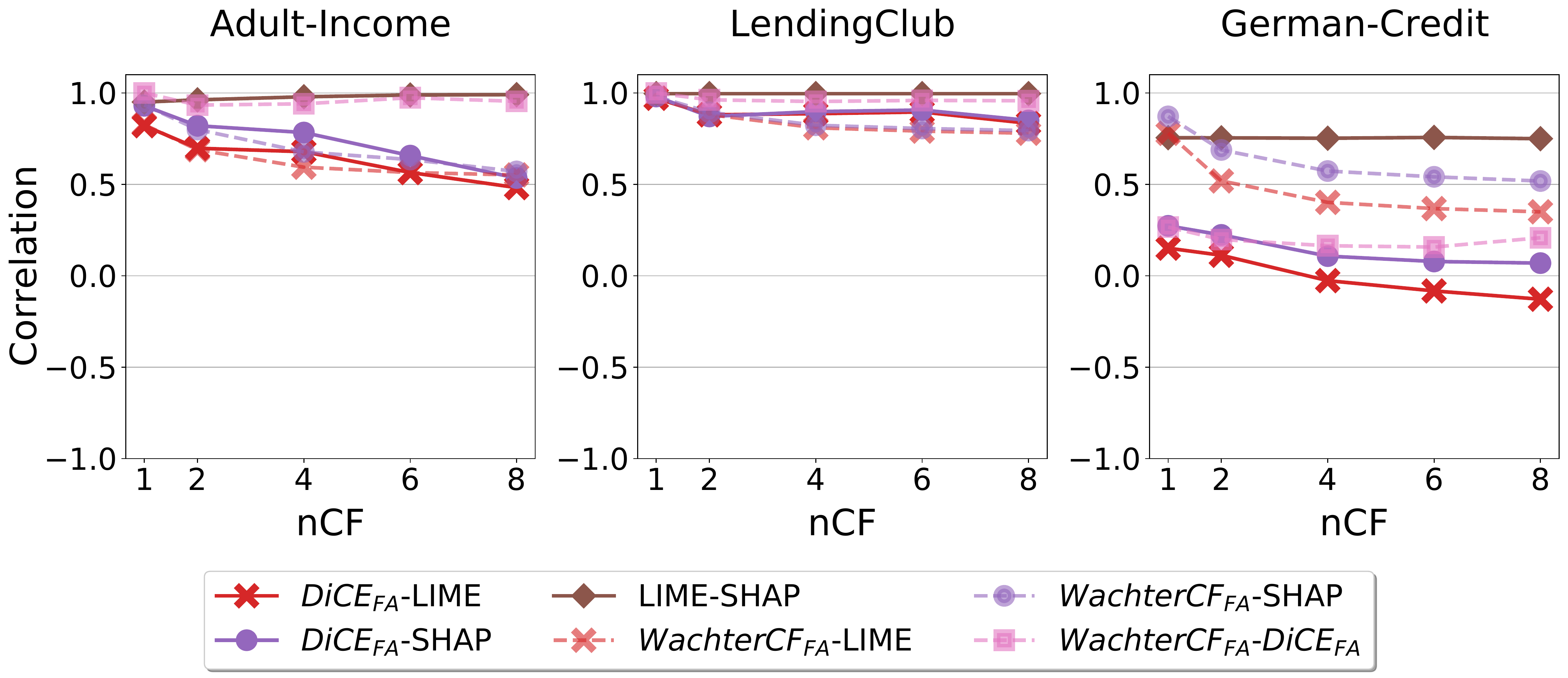}
    \caption{Correlation between feature importance scores.}
    \label{fig:feature_ranking_corr}
    \end{subfigure}
    \caption{In \figref{fig:feature_attribution_scores}, feature indexes on the $x$-axis are based on the ranking from \lime.
    Ranking from \shap mostly agrees with \lime, but less important features based on \lime can have high feature importance based on \wachtercffa and \dicefa.
    \figref{fig:feature_ranking_corr} shows the correlation of feature importance scores from different methods: \lime and \shap are more similar to each other than to \dicefa and \wachtercffa. In \german, the correlation with \dicefa can become negative as nCF grows.
    }
    \label{fig:feature_importance}
\end{figure*}

\para{Are important features sufficient?} 
Similar to necessity, we measure the sufficiency of top features from attribution-based methods by fixing the $k$-th most important feature and allowing \dice and \wachtercf to change the other features.
If the $k$-th most important feature is sufficient for the original prediction, we would expect a low success rate in generating valid CFs with the other features, and our sufficiency measure would take high values.

\figref{fig:all_data_gen_cfs_fix_topk_feats} shows the opposite. 
We find that the validity is close to 100\% till $nCF=8$ even {\em without} changing the $k$-th most important feature based on \lime or \shap in \adult and \german. This is the same as the validity (100\%) when changing all features, hence the sufficiency metric is near 0.
In comparison, for \lclub, while no change in the top-2 or top-3 does not affect the perfect validity, however, no change in the most important feature does decrease the validity when generating more than one CFs {\em using \dice}. 
This result again highlights the dominance of grade in \lclub.
However, even in this case, the sufficiency metric is  below 20\%.
Sufficiency results using \wachtercf are similarly low, except for \lclub when $nCF>1$. Here 
\wachtercf, with only random initialization and no explicit diversity loss formulation, could not generate multiple unique CFs  (without changing the most important features) for many inputs, and therefore the measured sufficiency is relatively higher. 
We also repeat the above analysis by fixing \textit{all} the top-$k$ features and get similarly low sufficiency results (see Suppl. \ref{sec:app-suff-nec} for more details). 



\para{Implications.}
These results qualify the interpretation of ``important'' features returned by common attribution methods like LIME or SHAP. Highly ranked features may often neither be necessary nor sufficient, and our results suggest that these properties become weaker for top-ranked features as the number of features in a dataset increases. In any practical scenario, hence, it is important to check whether necessity or sufficiency is desirable for an explanation. While feature importance rankings may be generally aligned with each feature's necessity, they can also deviate from this trend as we saw with LIME and \adult.
 In addition,  the results on \lclub indicate that  the method used to generate CFs matters too.  Defining the loss function with or without diversity  corresponds to different set of contexts on which necessity or sufficiency is estimated, which needs to be decided based on the application. Generally, whenever there are multiple kinds of attribution rankings to choose from,  these results demonstrate the value of using CFs to evaluate them.
\section{Feature Importance by CFs} \label{sec:all-feats}



As discussed in \secref{sec:methods}, counterfactual methods can not only evaluate, but also generate their own feature attribution rankings based on how often a feature is changed in the generated CFs.
In this section, we compare the feature importance scores from \dicefa and \wachtercffa to that from \lime and \shap, and investigate how they can provide additional, complementary information about a ML model.

\smallskip
\para{Correlation with \lime or \shap feature importance.} We start by examining how the importance scores from different methods vary for different features and datasets. 
\figref{fig:feature_attribution_scores} shows the average feature importance score across 200 random test instances when $\text{nCF}=4$. 
For \lime and \shap, we take the absolute value of feature importance score to indicate contribution.
\lime and \shap agree very well for \adult and \lclub.
While they mostly agree in \german, there are some bumps indicating disagreements.
In comparison, \dicefa and \wachtercffa are less similar to \lime than \shap.
This is especially salient in the high-dimensional \german dataset.
The features that are ranked 13th and 18th by \lime~--- the no. of existing credits a person holds at the bank and the no. of people being liable to provide maintenance for --- are the top two important features by \dicefa's scores. They are ranked 1st and 2nd, respectively, by \dicefa in 98\% of the test instances. Similarly, the 16th ranked feature by \lime, maximum credit amount, is the most important feature by \wachtercffa.



We then compute the Pearson correlation between these average feature importance scores derived with different explanation methods in Fig. \ref{fig:feature_ranking_corr} for different $\text{nCF}$. \lime and \shap agree on the feature importance  for all the three datasets, similar to what was observed in Fig. \ref{fig:feature_attribution_scores} at nCF=4. 
The correlation is especially strong for \adult and \lclub, each of which have only 8 features. 

Comparing CF-based and feature attribution methods, we find that they are well correlated in \lclub. 
This, again, can be attributed to the dominance of {\em grade}.
All methods choose to consider grade as an important feature.
In \adult, the correlation of CF-based methods with \shap and \lime decreases as $nCF$ increases.
This is not surprising since at higher $nCF$, while \dice changes diverse features of different importance levels (according to \lime or \shap) to get CFs, \wachtercf does so to a lesser extent with random initializations. 
For instance, in Fig. \ref{fig:feature_attribution_scores} at $nCF=4$, the feature that is ranked 6th on average by \lime, \textit{hours-per-week}, is changed by \wachtercf almost to the same extent as the top-3 features. Similarly, \dice varies this feature almost twice more than feature \textit{sex}, which is ranked 4th on average by LIME. 
Hence, we can expect that the average frequency of changing the most important feature would decrease with increasing $nCF$ and less important features would start to vary more (see \secref{sec:topk-feats}). 
By highlighting the less-important features as per \lime or \shap, \dicefa and \wachtercffa focuses on finding different subsets of \textit{necessary} features that can change the model output. In particular, even without a diversity loss, \wachtercffa varies less important features to get valid CFs.
 \lime and \shap instead tend to prefer \textit{sufficiency} of features in contributing to the original model output. 



This trend is amplified in \german dataset that has the highest number of features: correlation between \dicefa and \lime or \shap is below 0.25 for all values of $\text{nCF}$ and can even be negative as nCF increases.
We hypothesize that this is due to the number of features. 
\german has 20 features and in general with increasing feature set size, we find that \dice is able to generate CFs using less important features of \lime or \shap. Even though  \wachtercffa varies less important features as shown for nCF=4 in Fig. \ref{fig:feature_attribution_scores}, it  has a relatively moderate correlation with \lime/\shap. This implies that attribution-based and CF-based methods agree more when CFs are generated without diversity. Interestingly, the \wachtercffa  and \dicefa correlate less with each other than \wachtercffa correlates with \lime/\shap, indicating the multiple variations possible in generating CFs over high-dimensional data. 
Further, \lime and \shap also agree less in \german compared to other datasets, suggesting that datasets with few features such as \adult and \lclub may provide limited insights into understanding explanation methods in practice, especially as 
real-world datasets tend to be high-dimensional.

\para{Differences in feature ranking.}
Feature importance scores can be difficult to compare and interpret, therefore many visualization tools show the ranking of features based on importance. 
We perform a paired $t$-tests to test if there is a significant difference between rankings from different methods for the same feature. 
This analysis allows us to see the local differences in feature rankings beyond average feature importance score. For space reasons, we include the figures in our Supplementary Materials (see \ref{app:feat-ranks}, Figures \ref{fig:ttest-plots}, \ref{fig:ttest-plots-wachter}).

For most features across all datasets, we find that the feature rankings on individual inputs can be significantly different. 
In other words, the differences between explanation methods are magnified if we focus on feature ranking.
This is true even when comparing \lime and \shap, which otherwise show high positive correlation in average (global) feature importance score.
For instance, in \adult, \lime consistently ranks marital status and sex higher than \shap, while \shap tend to rank work class, race, and occupation higher.
Interestingly, they tend to agree on the ranking of continuous features, i.e., hours per week and age.
As expected, \lime and \dice provide different rankings for all features, while \shap and \dice differs in all except marital status.
Similar differences appear in feature rankings for \german and \lclub datasets.

\para{Implications.} 
Feature importance rankings by counterfactuals are quite different from attribution-based methods like \lime/\shap. In particular, they focus more on the less-important features from \lime/\shap and this trend accentuates as the number of features increases. A possible reason is that these explanations capture different theoretical notions such as  necessity and sufficiency, which is why \dicefa disagrees in its ranking on almost all features with \lime and \shap. 
This difference is not a critique to either method, rather an invitation to consider multiple explanation methods to complement each other. For example, in settings where necessity of features is important (e.g., algorithmic recourse for individuals), attribution rankings from CFs may be used in conjunction with the standard attribution-based methods.

At the same time, attributions from both kinds of explanation methods are sensitive to implementation details.  
While we expected significant differences between \dicefa and the two attribution-based methods based on global feature importance scores from Fig.~\ref{fig:feature_importance}, we also find significant differences between \lime and \shap on individual inputs, and between \dicefa and \wachtercffa on aggregate importances. 
In general, our results demonstrate the difficulty in building a single, ideal explanation method.

\section{Case Study: Hospital Admission}
To understand the complementarities between different explanation methods on a realistic dataset,
we 
 present a  case study 
 using a real-world hospital admission prediction problem with 222 features. 
Predicting patients who are likely to get admitted during emergency visits helps hospitals to better allocate their resources, provide appropriate medical interventions, and improving patient treatment rates \cite{haimovich2017discovery,desautels2016prediction,horng2017creating,levin2018machine,obermeyer2016predicting,moons2012risk,dugas2016electronic,arya2013decreasing}. Given the importance of the decision, it is critical that the predictions from an ML model be explainable to doctors in the emergency department. 
We leverage the dataset and models by Hong et al.~\cite{hong2018predicting} who use a variety of ML models including XGBoost and deep neural networks to predict hospital admission at the emergency department (ED). 

\begin{figure}[t]
    \centering
    \includegraphics[scale=0.3]{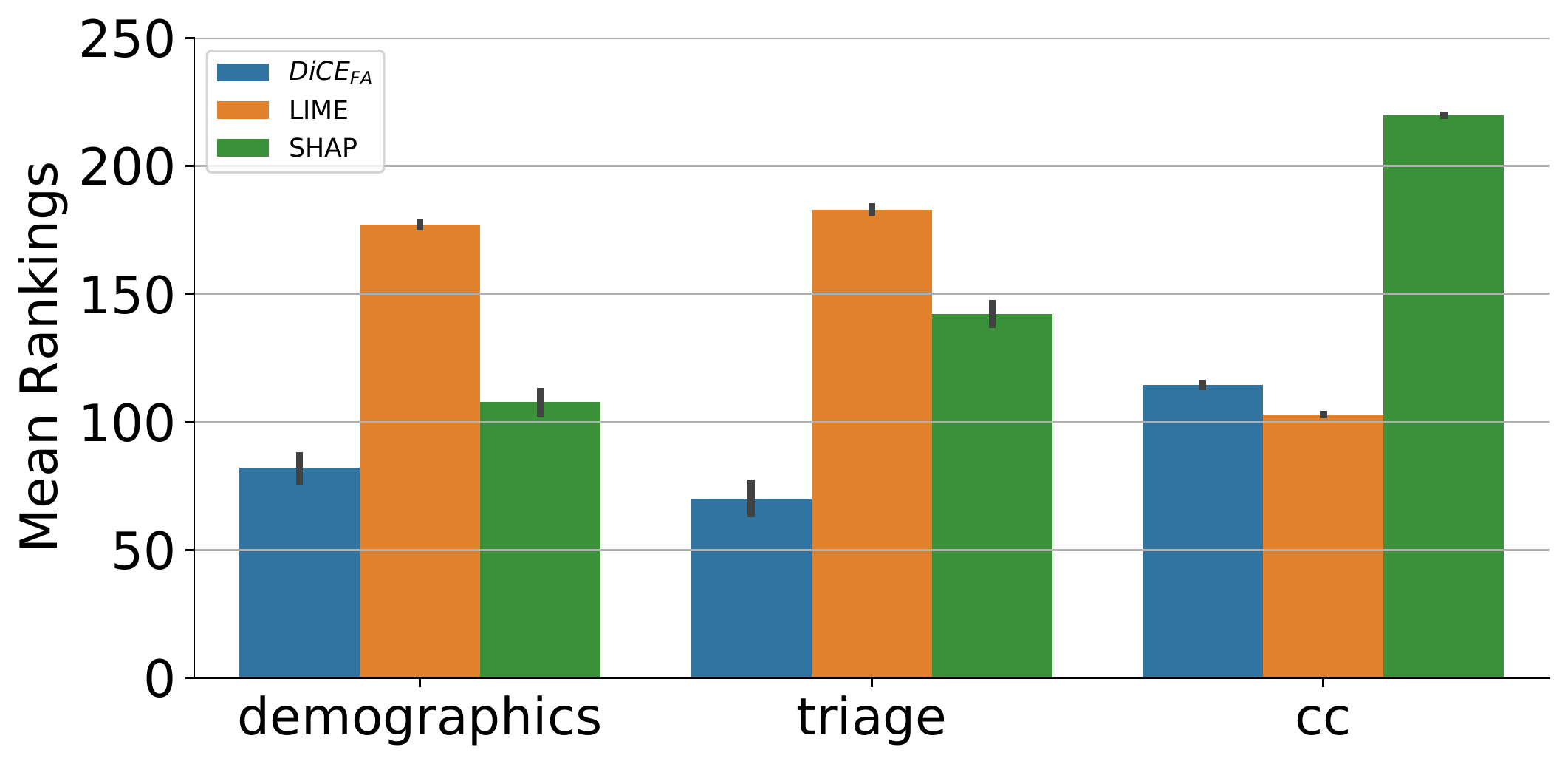}
    \caption{Mean rankings of different feature groups by \dicefa, \lime, and \shap.
    Lower rank $\Rightarrow$ higher importance.  
    }
    \label{fig:triage_mean_rankings}
\end{figure}

\para{Data and model training.} 
We use the ML model based on triage features, demographic features and chief complaints information from Hong et al.
Triage features consist of 13 variables to indicate the severity of ailments when a patient arrives at the ED. 
This model also uses 9 demographic features, including including race, gender, and religion, and 200 binary features indicating the presence of various chief complaints.
As a result, this dataset has many more features than \adult, \lclub, and \german.
We refer to this dataset as \triage.
We reproduce the deep neural network used by Hong et al. which has two hidden layers with 300 and 100 neurons respectively. The model achieves a precision and recall of 0.81 each and an AUC of 0.87 on the test set. We used a 50\% sample of the original data, consisting of 252K data points, for model training as the authors show that the accuracy  saturates beyond this point. We sample 200 instances from the test set over which we evaluate the attribution methods.


\para{In-depth look at the feature ranking.}
We start with the feature ranking produced by different methods to help familiarize with this real-world dataset.
We then replicate the experiments in \secref{sec:topk-feats} and \secref{sec:all-feats}. 
We focus on \dice in this comparison as \wachtercf can struggle to generate multiple unique valid CFs when $nCF>1$. 

\begin{figure}[t]
    \centering
    \begin{subfigure}{0.25\textwidth}
    \centering
        \includegraphics[width=\textwidth]{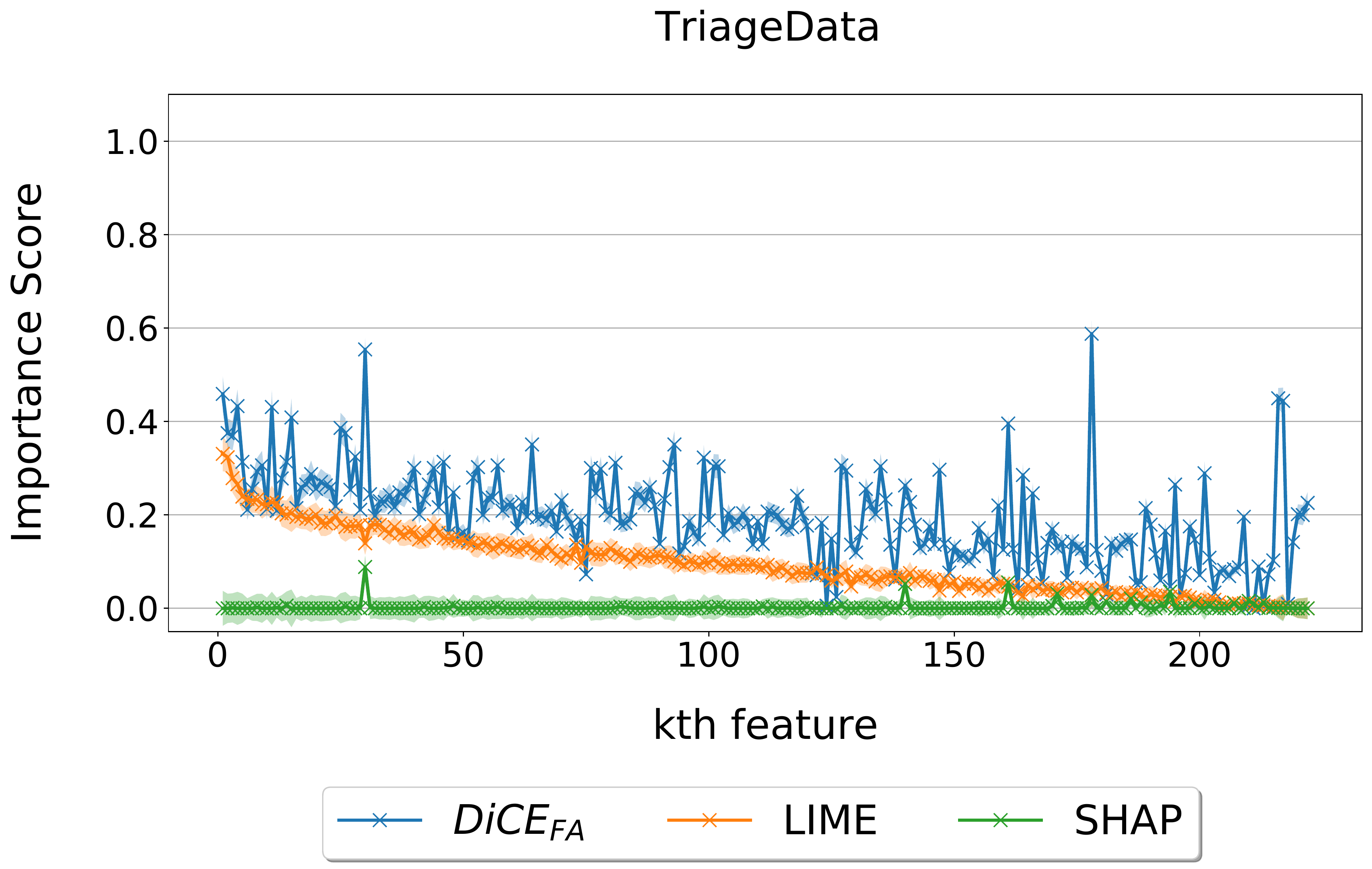}
    \caption{Average feature importance scores (nCF=4).}
        \label{fig:medical_feature_attribution_scores}   
    \end{subfigure}
    \hfill
    \begin{subfigure}{0.2\textwidth}
    \centering
    \includegraphics[width=\textwidth]{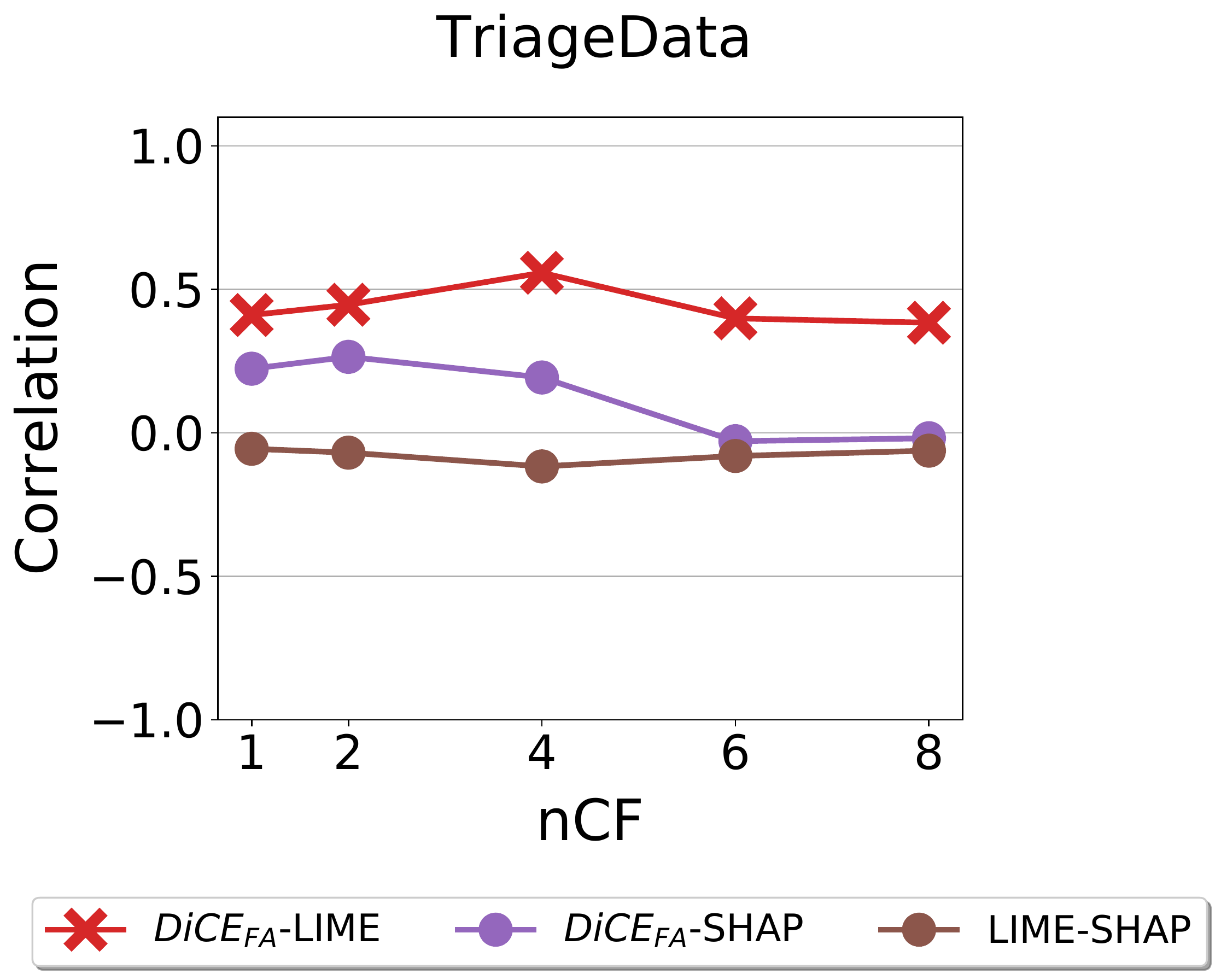}
    \caption{Correlation between feature importance scores.}
    \label{fig:medical_feature_ranking_corr}
    \end{subfigure}
    \caption{In \figref{fig:medical_feature_attribution_scores}, feature indexes in $x$-axis are based on ranking from \lime.
    \shap presents different outcomes from \lime, and their feature importance show much smaller variation than \dicefa.
    \figref{fig:medical_feature_ranking_corr}  compares feature importance score from different methods: the correlation between \lime and \shap is much weaker than in \figref{fig:feature_ranking_corr}.}
\end{figure}

We rank the features of \triage based on \dicefa,  \lime, and \shap using the same method as in \secref{sec:all-feats}.
\figref{fig:triage_mean_rankings} shows the distribution of mean rankings of different types of features in \triage according to our feature attribution methods.\footnote{We assign features the maximum of the ranks when there is a tie. \dicefa's and \lime's rankings are invariant to the treatment of ties whereas \shap's is. We choose the maximum to better distinguish different methods' rankings.}
This dataset has three category of features --- demographics, triage and chief complaints. 
We find that \shap ranks binary chief-complaints features much higher on average than \dicefa and \lime ($rank \propto \frac{1}{importance}$). Though \dicefa and \lime disagree on demographics and triage features rankings, they both have similar mean rankings on chief-complaints features which constitutes 90\% of the features. Hence, \dicefa and \lime has a relatively higher correlation (see \figref{fig:medical_feature_ranking_corr})  compared to any other pairs of methods. 

Furthermore, \dicefa considers demographics and triage features more important as compared to the chief-complaints features, since the former features have smaller rank ($<$80) on average. In contrast, \lime assigns them a larger rank. This has implications in fairness: when the ML model is evaluated based on \lime alone, the model would be seen as fair since chief-complaints features contribute more to the prediction on average. However, \dicefa and \shap show that demographic features can also be changed to alter a prediction, raising questions about making decisions based on sensitive features. 
Indeed, \citet{hong2018predicting} present a low-dimensional XGBoost model
by identifying features using information gain as the metric. They find 
that 5 out of 9 demographic details -- insurance status, marital status, employment status, race, and gender, and 6 out of 13 triage features are identified as important in their refined model. On the other hand, only 8 out of 200 chief-complaints features are found important.  
Note that these demographic details could be valid signals to use in health care; our main point is on the different interpretations of the same model by different methods.

\begin{figure}[t]
\centering
\begin{subfigure}[b]{0.45\textwidth}
   \includegraphics[width=1\linewidth]{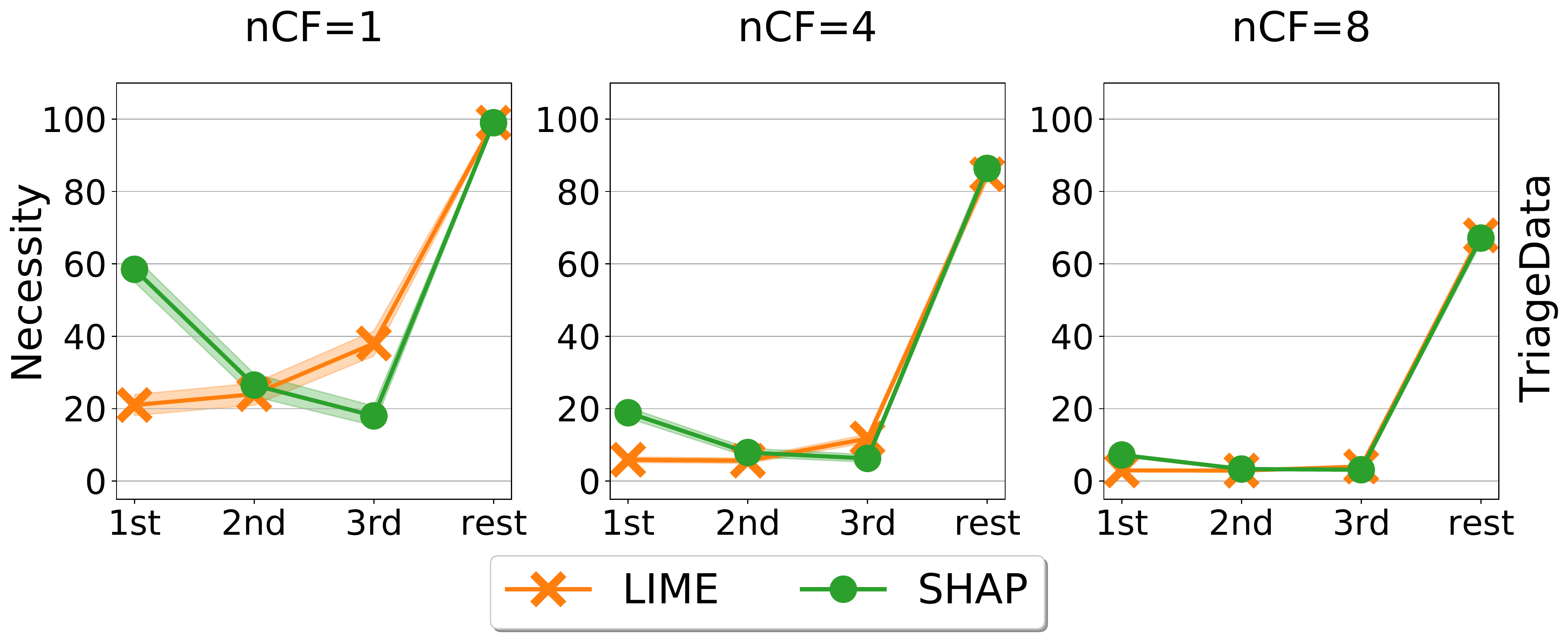}
   \label{fig:medical_gen_cfs_topk_feats_main}
\end{subfigure}\\
\vspace{-12pt}
\begin{subfigure}[b]{0.45\textwidth}
   \includegraphics[width=1\linewidth]{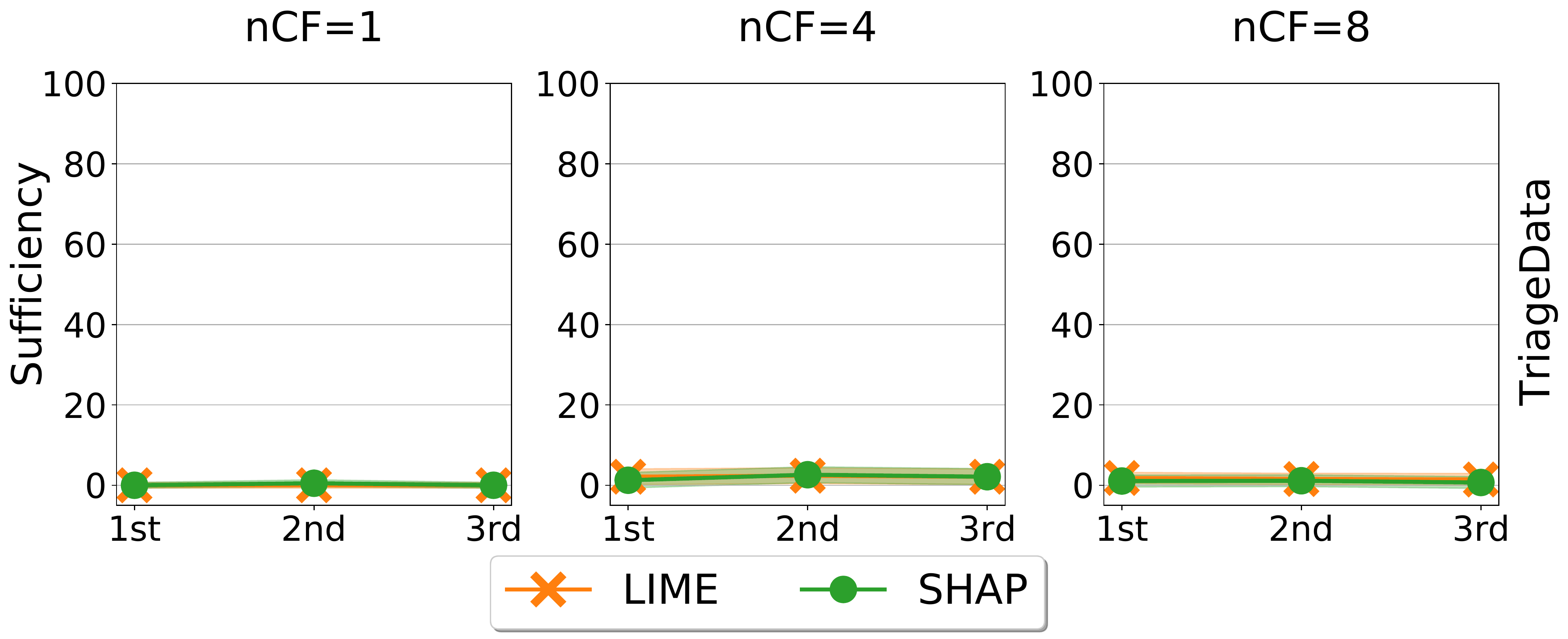}
   \label{fig:medical_gen_cfs_fix_topk_feats_main}
\end{subfigure}
\caption{\textit{Necessity} and \textit{Sufficiency} measures at a particular nCF, as defined in \secref{sec:necc-suff}, for the \triage data.}
\label{fig:medical_sufficiency_necessity}
\end{figure}

\para{Necessity and sufficiency.}
Next, we replicate the experiments from \secref{sec:topk-feats} for \triage
to understand the necessity and sufficiency of the important features of \lime and \shap in generating CFs.
The trend for \shap in \figref{fig:medical_sufficiency_necessity} is similar to what was observed in \figref{fig:all_data_gen_cfs_topk_feats}---
changing the more important features is more likely to generate valid CFs and hence higher necessity (green line).
However, in the case of \lime, we observe that the third important feature leads to more CFs, almost double than that of the first or second feature only. 
The reason is that in around 26\% of the test instances, \lime 
rates Emergency Severity Index (ESI) as the third most important feature. ESI
is a categorical feature indicating the level of severity assigned by the triage nurse \cite{hong2018predicting}. \dicefa considers this feature important to change the outcome prediction and ranks it among the  top-10 features for more than 60\% of the test instances. ESI is also one of the top-3 features by the information gain metric in the refined XGBoost model from Hong et al.

The sufficiency results (\figref{fig:medical_sufficiency_necessity}) are similar to \figref{fig:all_data_gen_cfs_fix_topk_feats}. Any of the top-3 features are \textit{not} sufficient
for generating CFs. At $\text{nCF}=1$, the same number of valid counterfactuals (100\%) can be generated while keeping the 1st, 2nd or the 3rd feature fixed, compared to the case when changing all features (and hence the sufficiency metric is near $0$). 
Similarly at $\text{nCF}=8$, the same number of valid counterfactuals (68\%) can be generated, irrespective of whether the top-k features are kept fixed or not. 
Note that the overall fraction of valid counterfactuals generated decreases as $\text{nCF}$ increases, indicating that it is harder  to generate diverse counterfactuals for this dataset. 
We expect the lack of sufficiency of top-ranked features to hold in many datasets, as the number of features increases. 

\para{Similarity between feature importance from different methods.}
\figref{fig:medical_feature_ranking_corr} shows the correlation of feature importance score derived from different methods.
Different from what was observed for other datasets in \figref{fig:feature_ranking_corr}, 
\lime and \shap have almost zero correlation between the feature rankings in \triage.  
This observation resonates with prior work demonstrating the instability and lack of robustness of these feature attribution methods, i.e., they can significantly differ when used to explain complex nonlinear models with high dimensional data \cite{zhang2019should,lai2019many,alvarez2018robustness,slack2019can}. 
In the case of \triage, the importance scores given by \lime and \shap are indeed very different for most of the features. For instance, \shap assigns close to zero weights for many binary ``chief-complaint'' features of \triage data in most of the test instances, while \lime assigns diverse importance scores. 
Fig. \ref{fig:medical_feature_attribution_scores} shows the absolute feature attribution scores of different methods at $nCF=4$ and it can be observed that \shap's scores are close to zero, on average, for most of the features.
Indeed, we find that the average entropy of the importance scores of \lime is 3.2 points higher than that of \shap on average. On the other other hand, the differences in entropy for \lclub, \adult, and \german were only 0.37, 0.48, and 0.84 respectively. 

In addition, while \dicefa agrees more with \shap than with \lime for other datasets (except \lclub where all methods agreed due to a dominating feature), here we obtain the reverse trend.  \dicefa has relatively weaker correlation with \shap in the case of \triage, echoing the difference observed for chief complaints in \figref{fig:triage_mean_rankings}. In particular, at $\text{nCF}=6$ and $\text{nCF}=8$, they both have no correlation on average feature rankings. At higher $\text{nCF}$, \dice varies more number of binary features most of which are assigned very low weights by \shap and hence the disagreement.


\para{Implications.} To summarize, we show how analyzing the feature attribution methods on a real-world problem highlights the complementarity and the differences in these methods. First, the highest ranked features by attribution-based methods like \lime are not sufficient, and are not always the most necessary for causing the original model output; more valid counterfactuals can be generated by varying a feature with larger rank compared to those with smaller rank.
Second, there are substantial differences in feature importance scores from the different methods, to the extent that they can completely change the interpretation of a model with respect to properties like fairness. Unlike the previous low-dimensional datasets, even \lime and \shap demonstrate substantial differences in global feature importance scores. 
\dicefa rankings somehow strike a balance between the two methods in importance:
\dicefa agrees with \shap on demographics features and with \lime on chief complaint features.
Finally,  similar to results in \secref{sec:all-feats}, \dicefa distributes feature importance more equally, especially for the features with larger rank from \lime and \shap.

\section{Concluding Discussion}

Our work represents the first attempt to unify explanation methods based on feature attribution and counterfactual generation.
We provide a framework based on actual causality to interpret these two approaches.
Through an empirical investigation on a variety of datasets,
we demonstrate intriguing similarities and differences between these methods.
Our results  show that it is not enough to focus on only the top features identified by feature attribution methods such as \lime and \shap.
They are neither sufficient nor necessary. 
Other features are (sometimes more) meaningful and can potentially provide actionable changes.

We also find significant differences in feature importance induced from different explanation methods. While feature importance induced from \dice and \wachtercf can be highly correlated with \lime and \shap on low-dimensional datasets such as \adult, 
 they become more different as the feature dimension grows.
Even in \german with 20 features, they can show no or even negative correlation when generating multiple CFs. Interestingly, we noticed differences even among methods of the same kind (\lime vs. \shap and \wachtercffa vs. \dicefa), indicating that more work is needed to understand the empirical properties of explanation methods on high-dimensional datasets. 

Our study highlights the importance of using different explanation methods and of future work to find which explanation methods are more appropriate for a given question.
There can be many valid questions that motivate a user to look for explanations \cite{liao2020questioning}.
Even for the specific question of which features are important,
the definition of importance can still vary, for example, actual causes vs. but-for causes.
It is important for our research community to avoid the one-size-fits-all temptation that there exists a uniquely best way to explain a model. 
Overall, while it is a significant challenge to leverage the complementarity of different explanation methods, we believe that the existence of different explanation methods provides exciting opportunities for combining these explanations.




\para{Acknowledgments.}
We thank anonymous reviewers for their helpful comments.
This work was supported in part by research awards from NSF IIS-2125116.

\bibliographystyle{ACM-Reference-Format}
\bibliography{references}

\appendix
\newpage

\appendix

\section{Supplementary materials}

\subsection{Explanation Scores: A Simple Example}
\label{app:exp-simple}

\begin{table}[h]
    \centering
    \begin{tabular}{l|lll}
         Method & $x_1$ & $x_2$ \\ 
         \hline
         \lime & 0.34 & 0.07 \\
         \shap (median BG) & 1.0 & 0.0 \\
         \shap (train data BG) & 0.69 & 0.28  \\
         \dicefa & 0.975 & 0.967 \\
         \wachtercffa & 1.0 & 0.975 \\
         \hline
    \end{tabular}
    \caption{Explaining model $y=I(0.45x_1+0.1x_2\geq 0.5)$ at an input point $(x_1=1, x_2=1, y=1)$. $x_1$ and $x_2$ are continuous features randomly sampled from an uniform distribution, $U(0,1)$. The second and third columns show an explanation method's score for $x_1$ and $x_2$ respectively. For \shap, the scores are shown for both median data and the entire training data as background (BG) sample in the second and third row respectively. Unlike attribution-based methods (LIME and SHAP), counterfactual-based methods (\dicefa and \wachtercffa) give almost equal importance to $x_2$ feature even though its coefficient in the target model is much smaller than $x_1$'s coefficient.}
    \label{tab:exp-simple-example}
\end{table}

\subsection{Choosing Counterfactual Explanation Methods}
\label{app:cf-algos}
We surveyed publicly available counterfactual explanation methods on GitHub which satisfy two criteria for our experiments: (a) support to generate counterfactuals using a subset of features, and (b) support to generate multiple counterfactuals. While few methods could be altered in theory to generate CFs using a feature subset \cite{karimi2020model,alibi,arya2019one,schleich2021geco}, we filter them out since it is not clear how to implement the same in practice without making significant changes to the original libraries. Similarly, we filter out those methods that do not explicitly support generating multiple CFs \cite{alibi,arya2019one}. 

Further, some libraries require substantial pre-processing to make comparison with other libraries for evaluation. For instance, while MACE \cite{karimi2020model} could generate multiple CFs, it requires extensive conversion to logic formulae to include any new ML model other than few standard models provided by the authors. Similarly, it is not clear how GeCo \cite{schleich2021geco}, written completely in Julia, could be altered to generate CFs with a feature subset (and how to use it to explain Python-based ML models and compare to other explanation methods which are mostly based in Python). DiCE \cite{mothilal2020explaining} and MOC \cite{dandl2020moc} are the only two libraries that directly satisfy both the aforementioned criteria. Further, the seminal counterfactual method by Wachter et al (\wachtercf) could also be easily implemented. Though \wachtercf, by default, provides only a single counterfactual, their optimization could be run with multiple random seeds to generate multiple counterfactuals simultaneously. Since we faced several compatibility issues such as transferring models between DiCE and MOC as these two libraries are based in Python and R respectively, we chose to use \dice and \wachtercf as our two counterfactual methods against the two feature attribution methods, \lime and \shap.  

\subsection{Datasets and Implementation Details}\label{app:impl-details}
We use three datasets.
\begin{itemize}[itemsep=0pt,leftmargin=*]
    \item \adult.  This dataset~\cite{adult} is based on the 1994 Census database and contains information like Age, Gender, Martial Status, Race, Education Level, Occupation, Work Class and Weekly Work Hours. It is available online as part of the UCI machine learning repository. The task is to determine if the income of a person would be higher than $\$50,000$ (1) or not (0). We process the dataset using techinques proposed by prior work~\cite{adult-helper} and obtain a total of 8 features.
    \item \lclub. Lending Club is a peer-to-peer lending company, which helps in linking borrowers and investors. We use the data about the loans from LendingClub for the duration 2007-2011 and use techniques proposed works~\cite{lendingclub-a, lendingclub-b, tan2017detecting} for processing the data. We arrive at 8 features, with the task to classify the payment of the loan by a person (1) versus no payment of the loan (0). 
    \item \german. German Credit~\cite{german} consists of various features like Credit Amount, Credit History, Savings, etc regarding people who took loans from a bank. We utilize all the features present in the dataset for the task of credit risk prediction, whether a person has good credit risk (1) or bad credit risk (0).
\end{itemize}

\para{Implementation Details.}
We trained ML models for different datasets in PyTorch and use the default parameters of LIME and DiCE in all our experiments unless specified otherwise. 
We use the same value of $\lambda_1$ for both \dice (Eqn.~\ref{eq:dice}) and \wachtercf (Eqn.~\ref{eq:wachtercf}).
For SHAP, we used its KernelExplainer interface with median value of features as background dataset. 
As SHAP's KernelExplainer is slow with a large background dataset, we used median instead. However, the choice of KernelExplainer and our background dataset setting can limit the strength of \shap 
\footnote{See issues 391 and 451 on SHAP's GitHub repository: https://github.com/slundberg/shap/issues},
and we leave further exploration of different configurations of SHAP to future work. 

Note that DiCE's hyperparameters for proximity and diversity in CFs are important. 
For instance, the diversity term enforces that different features change their values in different counterfactuals. Otherwise we may obtain multiple duplicate counterfactual examples that change the same feature. 
Results in the main paper are based on the default hyperparameters in \dice, but our results are robust to different choices of these hyperparameters (see Suppl. \ref{app:valid-stable}).

\subsection{Validity and Stability of \dice}
\label{app:valid-stable}

\begin{figure}[h]
    \centering
    \includegraphics[scale=0.19]{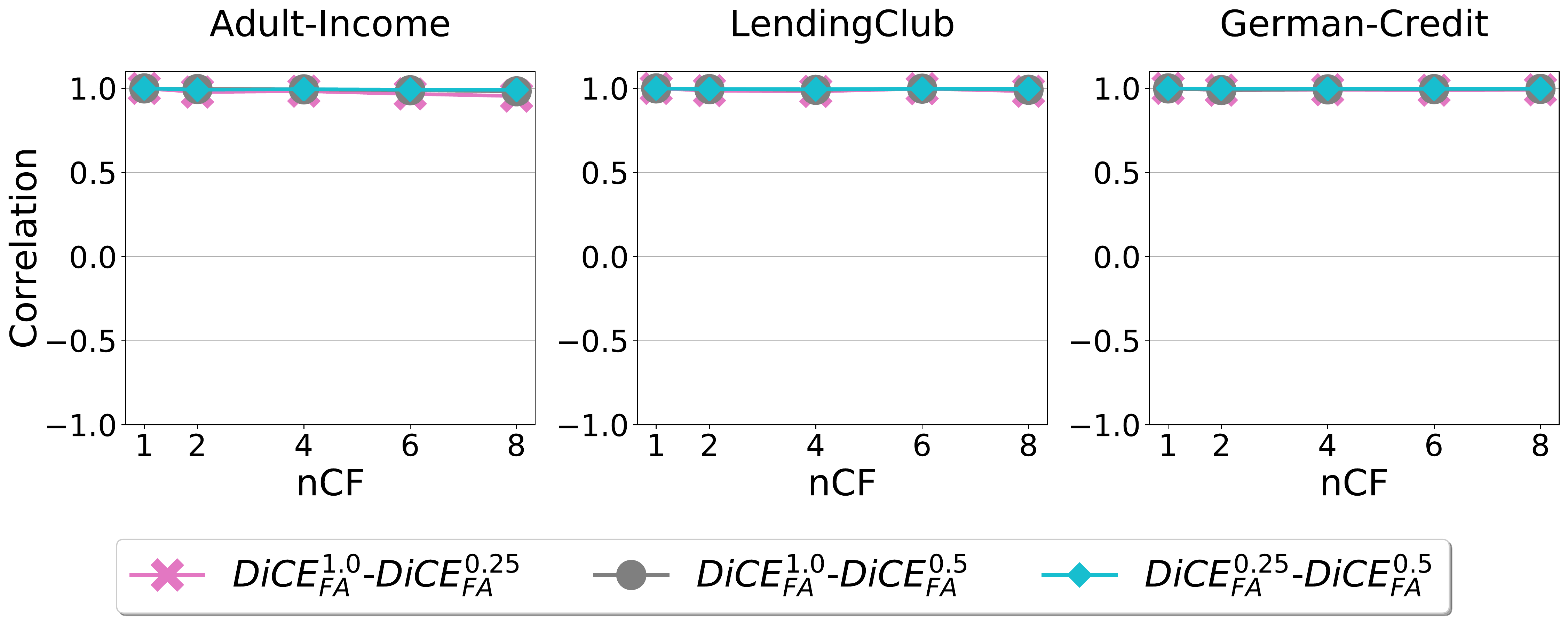}
    \caption{Correlation between different versions of \dicefa at different hyperparameters. The pink line corresponds to correlation between feature importance derived from \dice versions with 1.0 and 0.25 as \textit{diversity-weight} respectively. Similarly, the gray and blue lines correspond to 1.0 and 0.5, and 0.25 and 0.5 \textit{diversity-weights} respectively. All \dicefa methods exhibit high pairwise correlation ($>0.96$) on all datasets.
    }
    \label{fig:dice_mean_rankings}
\end{figure}

Table \ref{tab:gen_all_cfs_validity} shows the mean percentage validity of \dice with its default hyperparameters. \dice has two main hyperparamaters, namely \textit{proximity\_weight} and \textit{diversity\_weight}, controlling the closeness of counterfactuals to the test instance and the diversity of counterfactuals respectively. \textit{proximity\_weight} takes 0.5 and \textit{diversity\_weight} takes 1.0 as the default values respectively. These two parameters have an inherent trade-off \cite{mothilal2020explaining} and hence we change only the \textit{diversity\_weight} to examine the sensitivity of hyperparameters to the feature importance scores derived from \dicefa. Figure \ref{fig:dice_mean_rankings} shows the results. We find that \dicefa is not sensitive to these hyperparameters and different hyperparameter versions have a correlation of above 0.96 on all datasets.  

\begin{table}[h]
    \centering
    \resizebox{0.95\columnwidth}{!}{
    \begin{tabular}{l|lll}
         Data & avg \%valid CFs & \#instances \\ 
         \hline
         \adult & [96,99,98,98,98] & [192,196,188,184,185] \\
         \german & [100,100,100,100,100] & [199,198,199,199,198] \\
         \lclub & [100,100,100,100,100] & [200,200,200,200,200] \\
         \triage & [99,92,86,72,68] & [198,187,134,65,53] \\
         \hline

    \end{tabular}
    }
    \caption{The second column shows the mean percentage of unique and valid CFs found at each $nCF$ $\in$ \{1,2,4,6,8\} for different datasets given in the first column. The mean validity is computed over a random sample of 200 test instances for each dataset. The third column shows the number of test instances for which all the CFs found are unique and valid at different $nCF$.}
    \label{tab:gen_all_cfs_validity}
\end{table}

\subsection{Necessity and Sufficiency}
\label{sec:app-suff-nec}

\begin{figure*}[ht]
\centering
\begin{subfigure}{.48\textwidth}
  \centering
  \includegraphics[width=.95\textwidth]{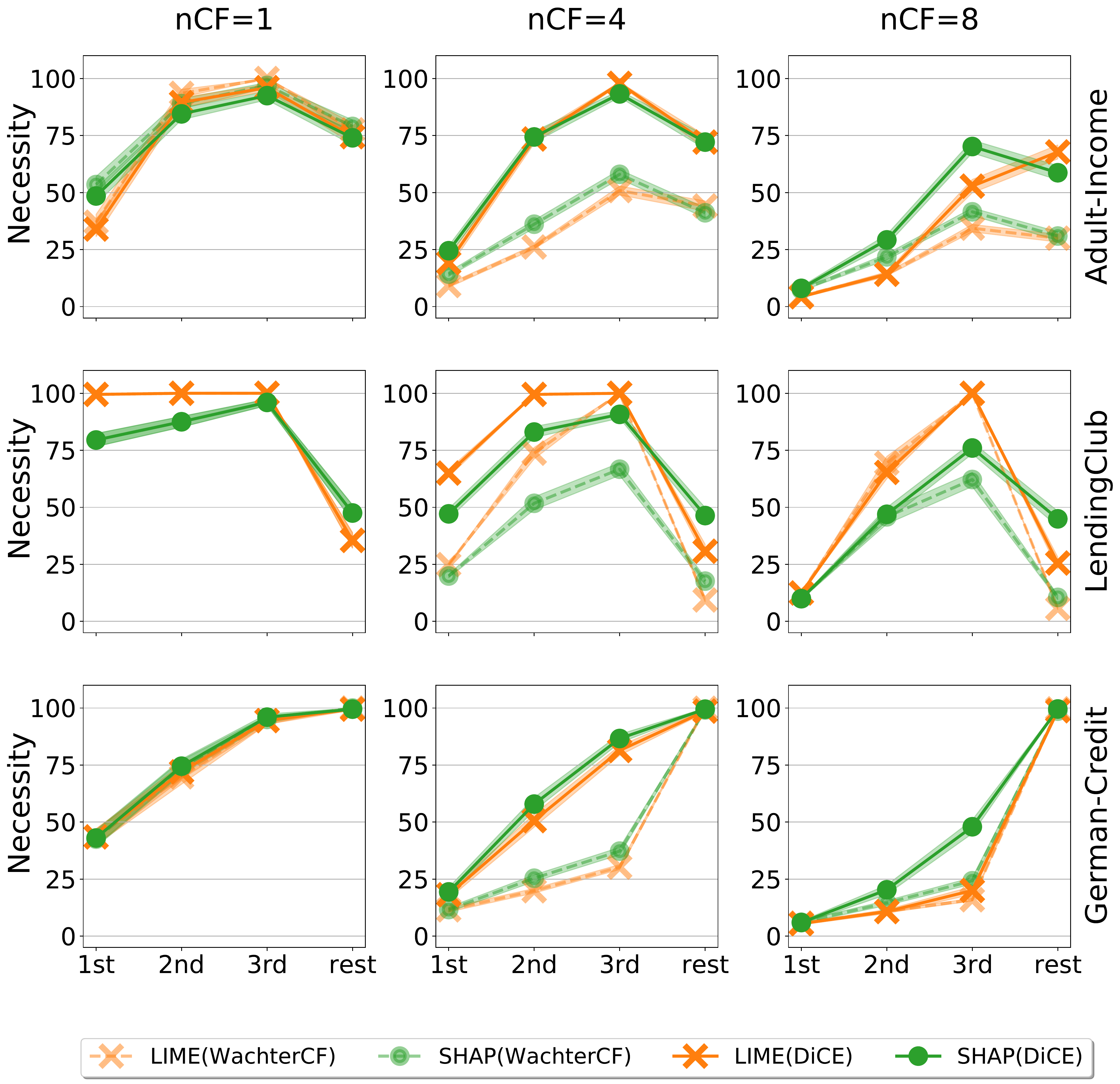}
  \caption{Necessity}
  \label{fig:all_data_gen_cfs_until_topk_feats}
\end{subfigure}
\hfill
\begin{subfigure}{.48\textwidth}
  \centering
  \includegraphics[width=.95\textwidth]{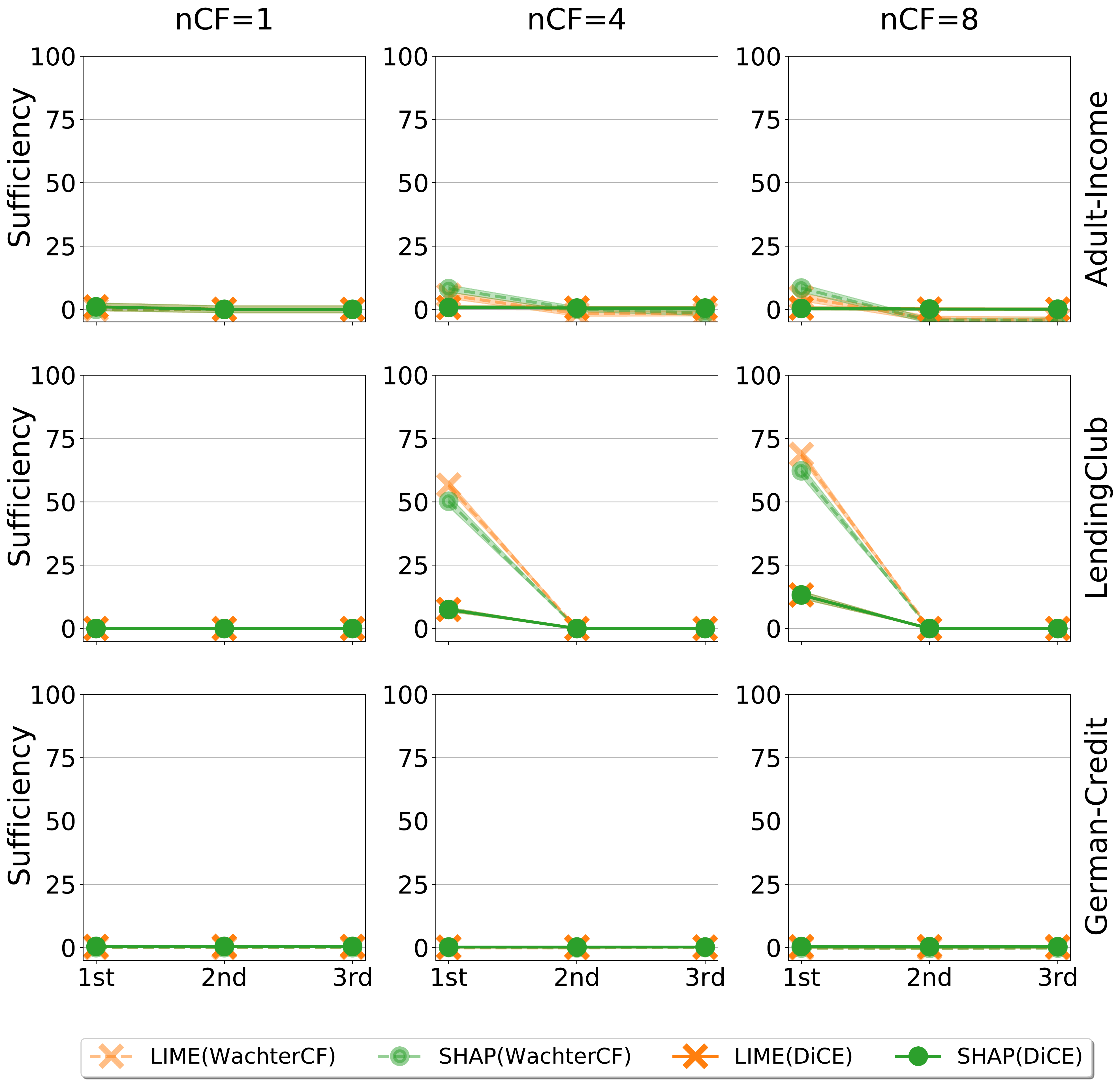}
  \caption{Sufficiency}
  \label{fig:all_data_gen_cfs_fix_until_topk_feats}
\end{subfigure}%
\caption{The $y$-axis represents the \textit{necessity} and \textit{sufficiency} measures at a particular $\text{nCF}$, as defined in \secref{sec:necc-suff}.
Fig. (a) shows the results when we are only allowed to change \textit{until} the $k$-th most important features ($k=1,2,3$) or the other features,
while Fig. (b) shows the results when we fix \textit{until} $k$-th most important features ($k=1,2,3$) but are allowed to change other features.
}
\label{fig:appendix-nec-suff-until}
\end{figure*}

Figure~\ref{fig:appendix-nec-suff-until} shows the necessity and sufficiency metrics when we allow \textit{all} features upto top-$k$ features to change (for necessity) or remain fixed (sufficiency). 
Necessity increases for the top-$k$ features, but sufficiency remains identical to the setting in the main paper. 

\subsection{Differences in Feature Rankings}
\label{app:feat-ranks}

\begin{figure*}[ht]
\centering
\subfloat[][Adult Dataset]{
  \includegraphics[width=0.30\linewidth, ]{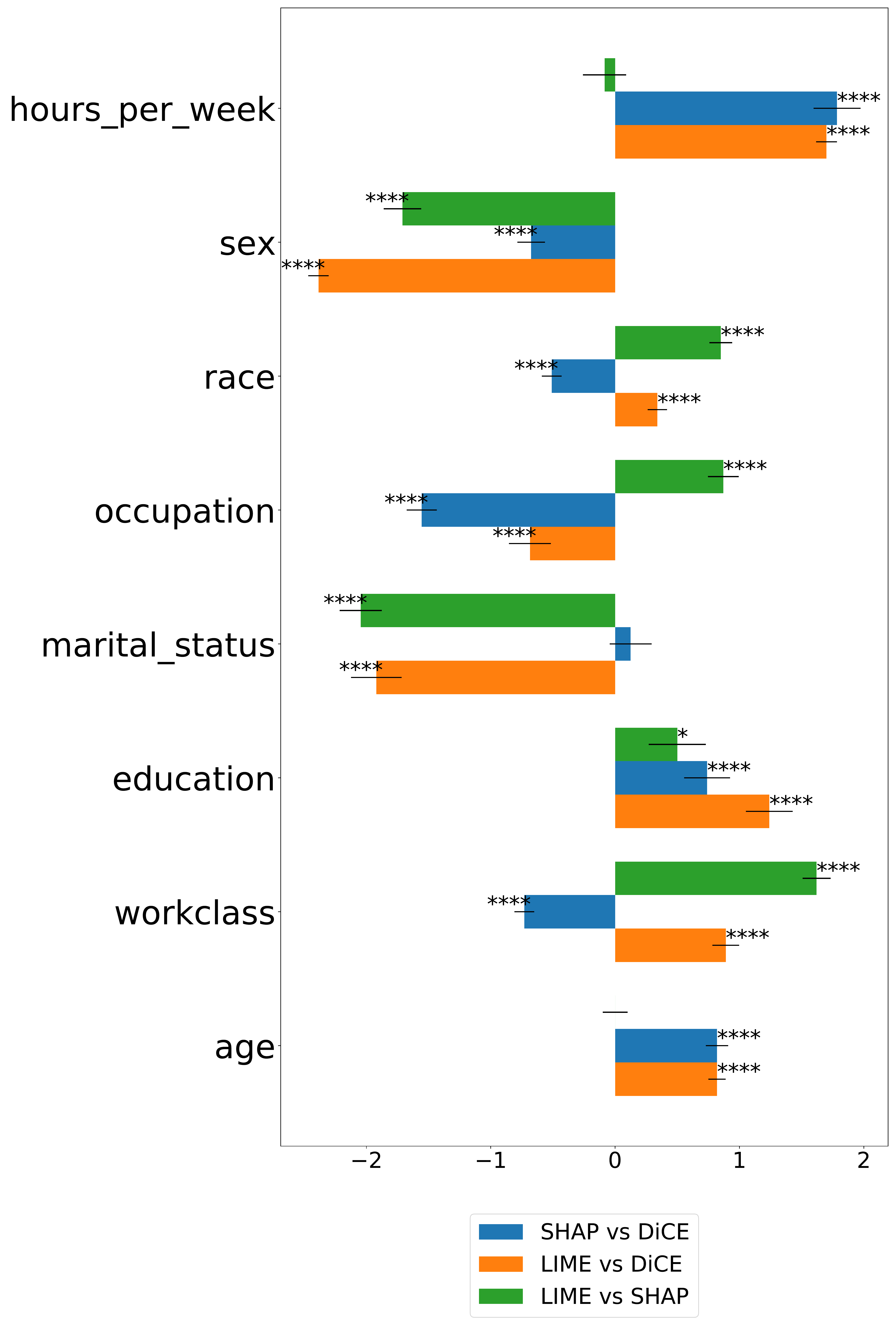}
  \label{fig:adult-ttest}
  }
  \quad
\subfloat[][Lending Club Dataset]{
  \includegraphics[width=0.30\linewidth]{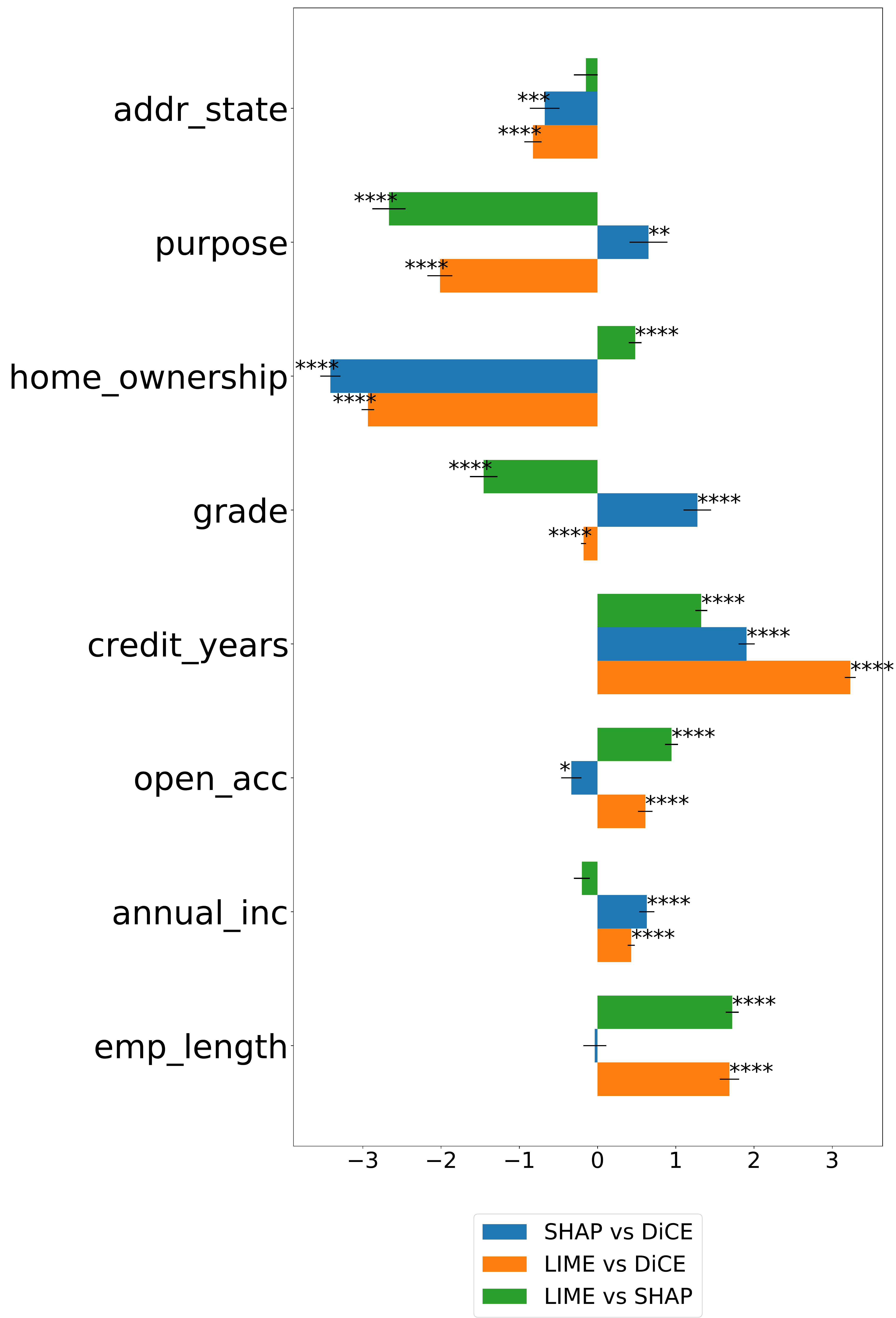}
  \label{fig:lending-ttest}
  }
  \quad
\subfloat[][German Credit Dataset]{
  \includegraphics[width=0.30\linewidth]{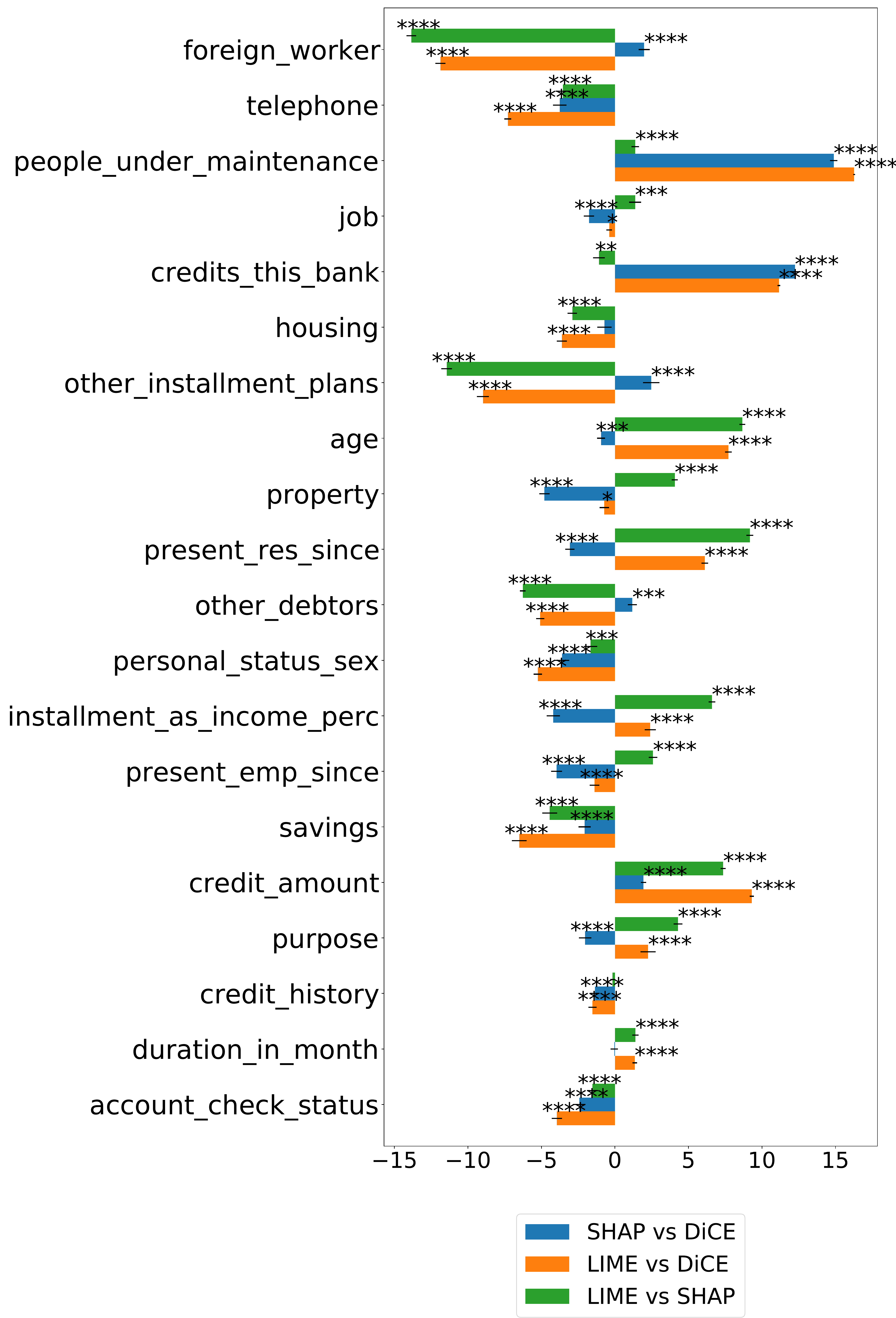}
  \label{fig:german-ttest}
  }   
 \caption{Correlation between the importance ranking of a feature across instances by \lime, \shap, and \dice. The x-axis denotes the mean difference in the rankings for each feature over all the test inputs. Stars denote significance levels using p-values (\textsuperscript{****}: $p<10^{-4}$, 
 \textsuperscript{***}: $p<10^{-3}$, 
  \textsuperscript{**}: $p<10^{-2}$, 
  \textsuperscript{*}: $p<5*10^{-2}$) 
  }
\label{fig:ttest-plots}
\end{figure*}

\begin{figure*}[ht]
\centering

\subfloat[][Adult Dataset]{
  \includegraphics[width=0.30\linewidth, ]{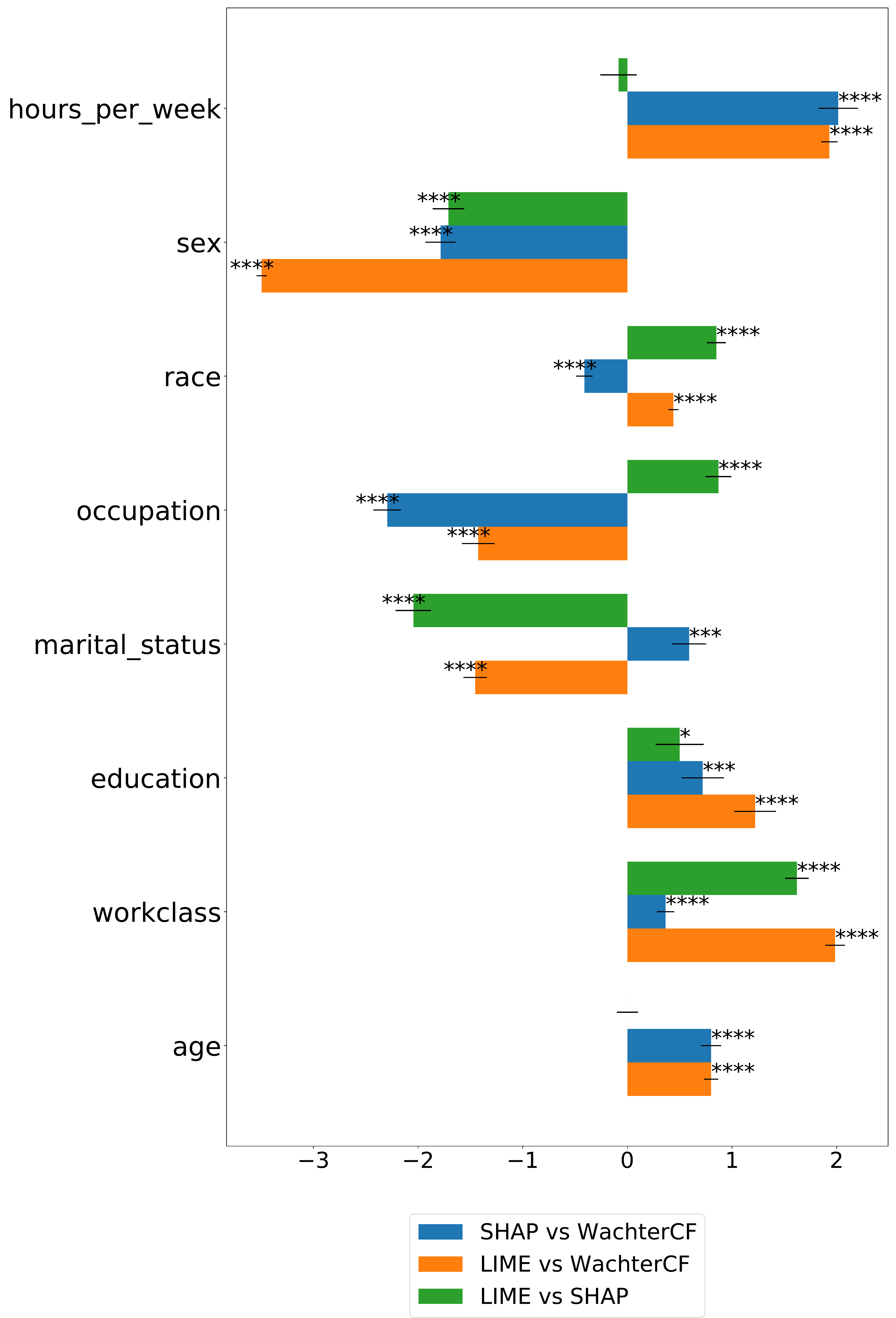}
  \label{fig:adult-wachter-ttest}
  }
  \quad
\subfloat[][Lending Club Dataset]{
  \includegraphics[width=0.30\linewidth]{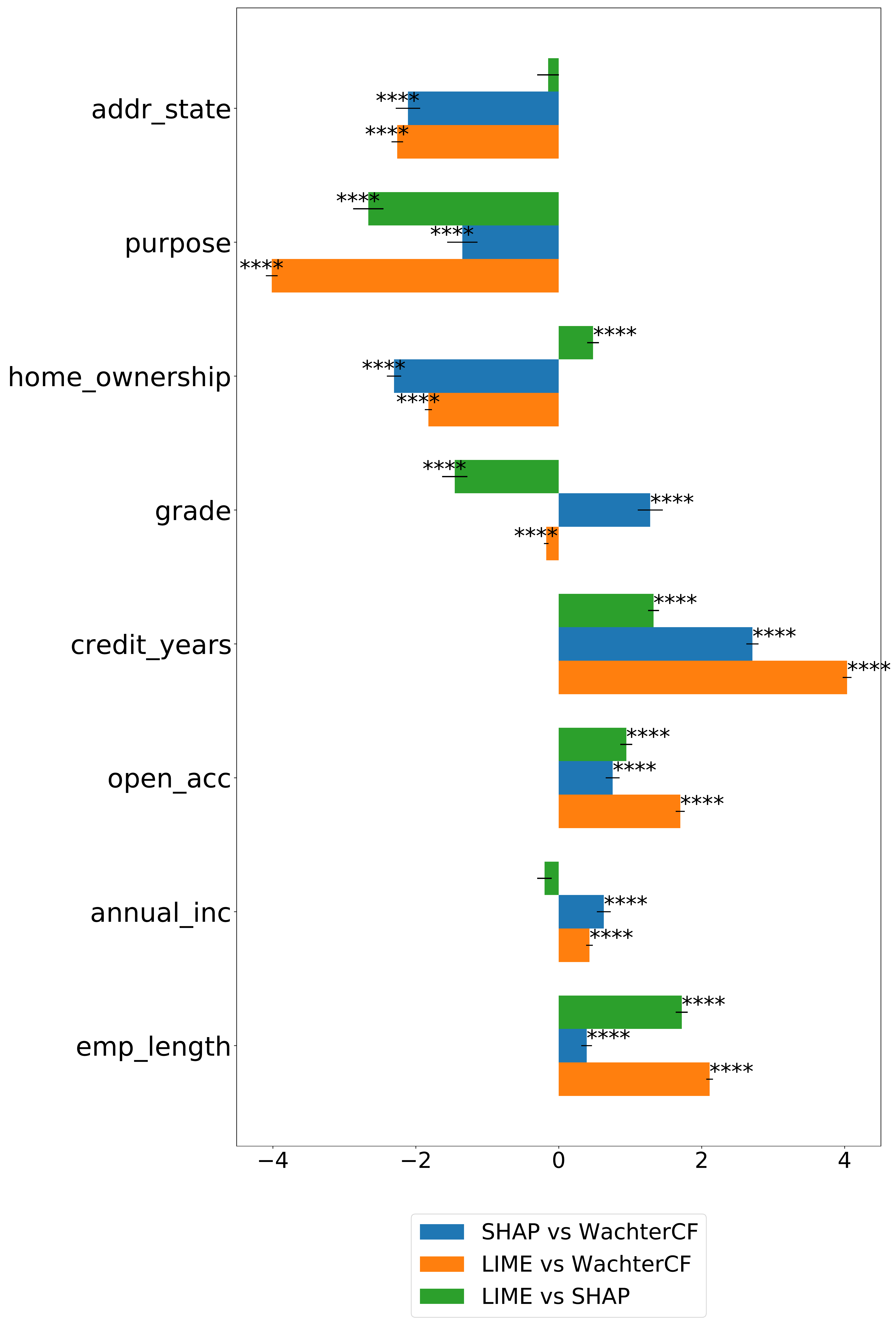}
  \label{fig:lending-wachter-ttest}
  }
  \quad
\subfloat[][German Credit Dataset]{
  \includegraphics[width=0.30\linewidth]{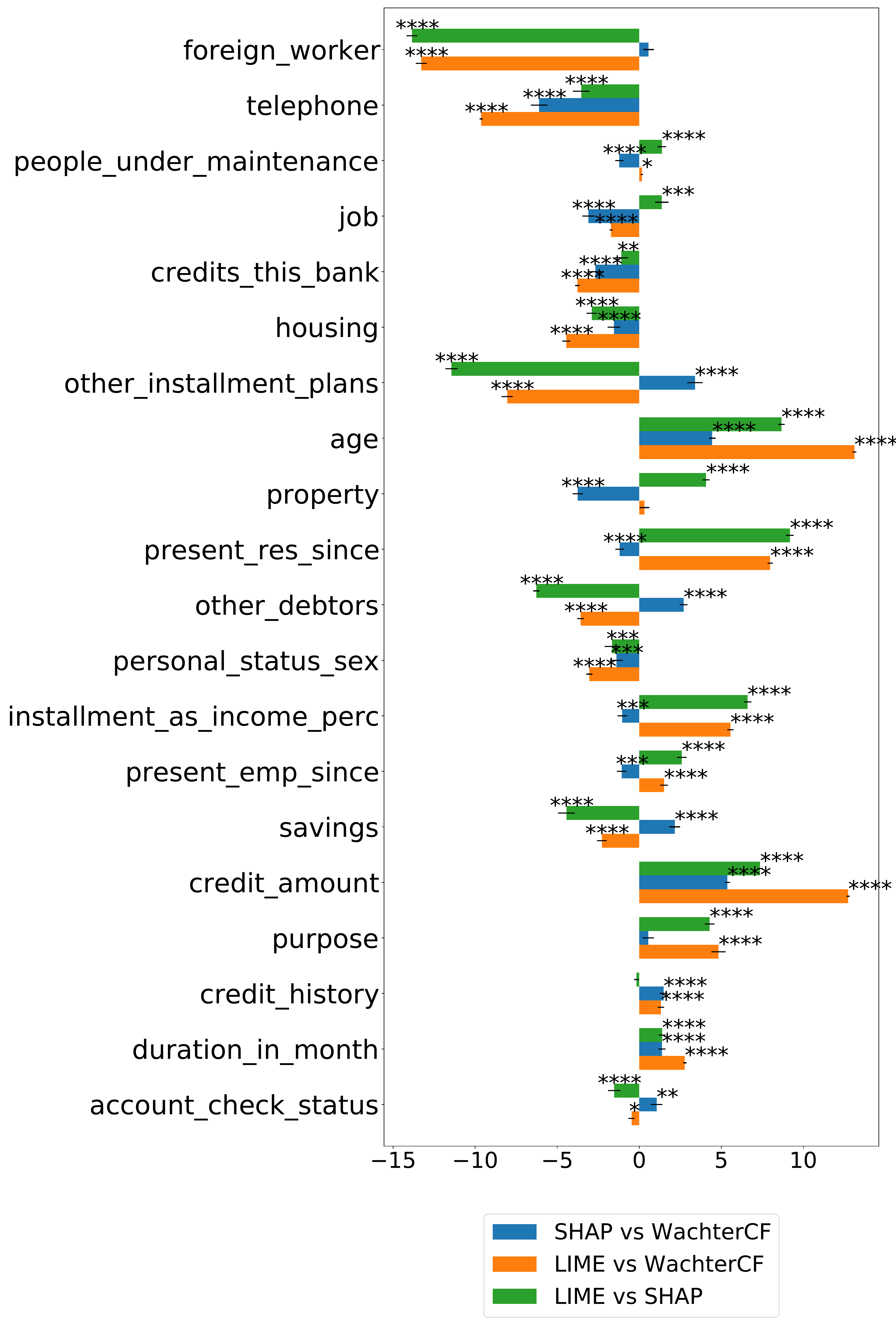}
  \label{fig:german-watchter-ttest}
  }   
 \caption{Correlation between the importance ranking of a feature across instances by \lime, \shap, and \wachtercf. The x-axis denotes the mean difference in the rankings for each feature over all the test inputs. Stars denote significance levels using p-values (\textsuperscript{****}: $p<10^{-4}$, 
 \textsuperscript{***}: $p<10^{-3}$, 
  \textsuperscript{**}: $p<10^{-2}$, 
  \textsuperscript{*}: $p<5*10^{-2}$) }
\label{fig:ttest-plots-wachter}
\end{figure*}

\end{document}
\endinput